\documentclass[11pt,letterpaper]{iffyuan}

\usepackage[all]{hypcap}
\usepackage{algorithm}
\usepackage{algorithmicx}
\usepackage{algpseudocode}
\usepackage{float}
\usepackage{mathtools}
\usepackage{nicefrac}
\usepackage{subcaption}
\usepackage{wrapfig}
\usepackage{xspace}
\usepackage{tikz}
\usepackage{array}

\title{ForceFlow: Learning to Feel and Act via Contact-Driven Flow Matching}

\author[1]{Shuoheng Zhang\textsuperscript{\faGem}}
\author[1]{Yifu Yuan\textsuperscript{\faGem,\faEnvelope\,}}
\author[1]{Hongyao Tang}
\author[1]{Yan Zheng}
\author[3]{Qiaojun Yu}
\author[1]{Pengyi Li}
\author[2]{Guowei Huang}
\author[2]{Helong Huang}
\author[2]{Xingyue Quan}
\author[1]{Jianye Hao\textsuperscript{\faEnvelope\,}}

\affiliation[1]{Tianjin University}
\affiliation[2]{Huawei Noah's Ark Lab}
\affiliation[3]{Shanghai AI Lab}

\contribution[\faGem]{Equal contribution}
\contribution[\faEnvelope]{Corresponding Author (\href{mailto:yuanyf@tju.edu.cn}{yuanyf@tju.edu.cn}, \href{mailto:jianye.hao@tju.edu.cn}{jianye.hao@tju.edu.cn})}

\setteamlogo{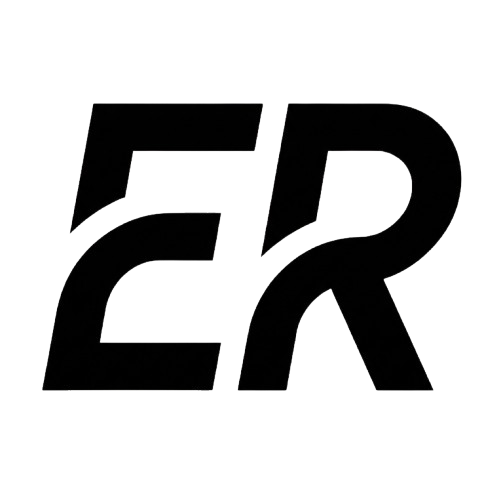}

\abstract{%
Existing imitation learning methods enable robots to interact autonomously with the physical environment. However, contact-rich manipulation tasks remain a significant challenge due to complex contact dynamics that demand high-precision force feedback and control. Although recent efforts have attempted to integrate force/torque sensing into policies, how to build a simple yet effective framework that achieves robust generalization under multimodal observations remains an open question. In this paper, we propose \textbf{ForceFlow}, a force-aware reactive framework built upon flow matching. For contact-stage policy design, we investigate force signal fusion mechanisms and adopt an asymmetric multimodal fusion architecture that treats force as a global regulatory signal, combined with a joint prediction paradigm that enhances the policy's understanding of instantaneous force and historical information, thereby achieving deep coupling between force and motion. For task-level hierarchical decomposition, we divide manipulation into a vision-dominant approach stage (VLM-based pointing for target localization) and a touch-dominant interaction stage (force-driven contact execution), with a Vision-to-Force~(V2F) handover mechanism that explicitly decouples spatial generalization from contact regulation. Experimental results across six real-world contact-rich tasks demonstrate that ForceFlow achieves a 37\% success rate improvement over the strong baseline ForceVLA while maintaining significantly lower cost. Moreover, ForceFlow exhibits accurate force signal prediction and demonstrates superior performance in contact force self-regulation and zero-shot out-of-distribution~(OOD) generalization.
}

\metadata[\faGlobe\ Project]{\href{https://jokeresc.github.io/ForceFlow-page}{https://jokeresc.github.io/ForceFlow-page}}
\metadata[\raisebox{-1.5pt}{\includegraphics[height=1.03em]{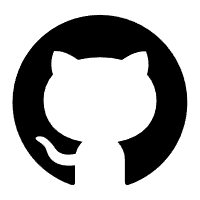}}\ Code]{\href{https://github.com/JokerESC/ForceFlow}{https://github.com/JokerESC/ForceFlow}}
\metadata[\raisebox{-1.5pt}{\includegraphics[height=0.96em]{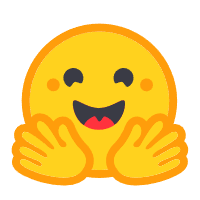}}\ Datasets]{\href{https://huggingface.co/datasets/JokerESC/ForceFlow}{https://huggingface.co/datasets/JokerESC/ForceFlow}}
\metadata[\faEnvelope\ Contact]{\href{mailto:zsh_2024244119@tju.edu.cn}{zsh\_2024244119@tju.edu.cn}}

\begin{document}

\maketitle

\section{Introduction}

In contact-rich manipulation, humans follow a natural staged perception process: vision is first used to localize the target and guide the hand toward it, while perceptual dominance shifts from vision to force and touch once physical contact is established~\cite{vital}. For tasks such as precision assembly, surface wiping, and peg-in-hole insertion~\cite{foar}, successful execution requires not only reaching the correct interaction region, but also regulating contact forces through rapid local adjustment. Enabling robots to achieve a similar seeing-to-feeling modal transition is a core problem in general-purpose embodied manipulation~\cite{tavla,forcevla,ma2025ddcg}.

Recent progress in imitation learning has enabled robots to acquire diverse manipulation skills from expert demonstrations~\cite{act,diffusion_policy,openvla}. However, these methods remain fundamentally vision-centric. Although visual observations are effective for macro-level positioning, they are often insufficient to independently support fine contact regulation due to occlusion, limited spatial resolution, and weak sensitivity to subtle contact events at the interaction interface~\cite{forcemimic}. Consequently, many vision-dominant policies can move the end-effector near the target, yet still struggle to reliably detect misalignment, regulate compliance, or adapt to changes in stiffness and friction during physical interaction.

Recent efforts have begun to incorporate force/torque signals into imitation learning frameworks~\cite{tavla,forcemimic,vital}, but two key challenges remain. The first is multimodal fusion under contact dynamics: force signals are low-dimensional, high-frequency, and strongly temporally structured, making them easily overshadowed by high-dimensional visual features during end-to-end training, which prevents policies from fully exploiting force information. The second is the dual generalization requirement: contact-rich manipulation simultaneously demands \emph{physical interaction generalization} (stable execution under unseen stiffness, friction, and force responses) and \emph{spatial generalization} (reaching the correct interaction region despite significant visual or positional shifts). Existing methods typically do not explicitly separate these two difficulties, leading to brittle performance when either physical properties or workspace layout changes.

\begin{figure}[t]
  \vspace{-1.5cm}
  \centering
  \includegraphics[width=\textwidth]{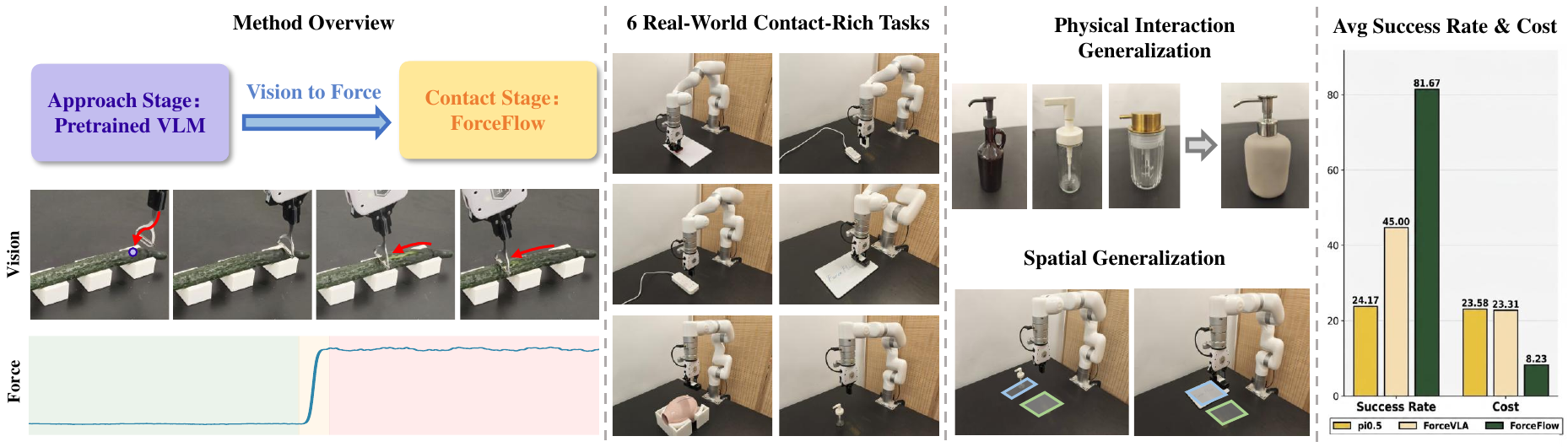}
  \caption{\textbf{ForceFlow integrates V2F handover mechanism with flow matching to achieve precise robotic manipulation in contact-rich tasks, demonstrating significantly higher success rates and superior OOD generalization compared to baselines.}}
  \label{fig:main_plot}
  \vspace{-0.5cm}
\end{figure}

To address these challenges, we propose \textbf{ForceFlow}, a force-aware reactive control framework built on Flow Matching~\citep{flowmatching}. We choose Flow Matching as the policy backbone because its deterministic ODE path generation provides lower inference latency and more stable trajectory output compared with diffusion sampling, which is critical for closed-loop contact control that must respond to high-frequency force feedback in real time. At the policy level, ForceFlow integrates temporal force history as a global regulatory signal into the network through asymmetric multimodal fusion, while retaining vision as a selective spatial reference, thereby effectively preventing force signals from being overshadowed by visual features. The policy jointly predicts actions and next-step contact force, encouraging the model to internalize the coupling between force and motion and enabling proactive force regulation. At the task level, we further introduce a \textbf{Vision-to-Force (V2F)} handover mechanism that divides fine manipulation into a vision-dominant approach stage and a force-dominant interaction stage. In the former, a Vision-Language Model localizes the target through a pointing mechanism; in the latter, once the end-effector reaches the VLM-derived 3D approach waypoint, the ForceFlow policy takes over for contact-rich execution. This design explicitly decouples spatial generalization from contact regulation, reflecting the natural transition from seeing to feeling in contact-rich manipulation. We systematically evaluate ForceFlow on six real-world contact-rich tasks spanning short-horizon contact establishment and continuous contact regulation. ForceFlow achieves an average success rate of 81.67\%, outperforming the strong baseline ForceVLA by 37\% while maintaining significantly lower contact force cost. Furthermore, ForceFlow exhibits stronger robustness under both unseen physical property variations and spatially out-of-distribution settings.

Our contributions are threefold: (1) We propose ForceFlow, a flow-matching policy for contact-rich manipulation that integrates temporal force history through asymmetric multimodal fusion and jointly predicts motion and next-step contact force. (2) We introduce the V2F handover mechanism that separates vision-guided spatial localization from force-driven contact execution, following a staged transition from vision dominance to force dominance. (3) We validate the framework on six real-world benchmark tasks, demonstrating clear gains over strong baselines including ForceVLA and $\pi_{0.5}$ in task success rate, contact force fidelity, and robustness to both physical and spatial distribution shifts.

\section{Related Work}

\paragraph{Imitation Learning for Robotics}
Imitation Learning (IL)~\citep{IL1} acquires manipulation skills from expert demonstrations. Early Behavioral Cloning (BC)~\citep{IL2} struggles in complex long-horizon tasks, prompting the adoption of generative models. Methods like ACT~\citep{act} and Diffusion Policy~\cite{diffusion_policy} enable long-horizon smooth trajectory generation. More recently, Vision-Language-Action (VLA) models (e.g., OpenVLA~\citep{openvla}, $\pi_{0.5}$~\citep{pi0.5}, and GROOT N1~\citep{grootn1}) leverage large-scale pre-training for open-world generalization. Despite these advances, these state-of-the-art methods remain inherently vision-centric, treating manipulation primarily as a kinematic trajectory generation problem. By neglecting high-frequency force-torque dynamics, they often exhibit brittleness in contact-rich scenarios where visual feedback becomes unreliable, and success is dictated by the precise regulation of contact forces. To address this issue, we propose ForceFlow, which employs a flow matching architecture supplemented by historical force feedback and contact force prediction to ensure accurate force regulation in contact-rich tasks.

\paragraph{Contact-Rich Manipulation}

Contact-rich tasks require continuous force regulation~\citep{zhao2026fd,li2026forcevla2}, motivating the integration of force or tactile signals into policy learning. Existing approaches typically introduce force as additional observations~\citep{feeltheforce,wu2025tacdiffusion} or infer it from vision and proprioception~\citep{forcemimic}. Other works focus on multimodal fusion via contact-aware gating~\citep{foar}, residual refinement~\citep{vital}, or hierarchical modeling~\citep{rdp}. Several recent models~\citep{forcevla,omnivtla,hao2025tla,huang2025tactile,tavla} further incorporate force/torque information into VLA-style frameworks. However, a common bottleneck persists: high-dimensional visual features tend to dominate end-to-end optimization, causing low-dimensional yet temporally rich force signals to be underutilized, a phenomenon known as \emph{modal masking}~\citep{multimodal1,multimodal2,conditioning}. Moreover, the diversity of sensor configurations across prior work complicates fair comparison. We therefore adopt the same observation setting as ForceVLA (multi-view RGB, end-effector pose, and 6D force/torque) and focus on suppressing modal masking through asymmetric fusion.

\section{The ForceFlow Framework}

To mitigate the critical issue where high-dimensional visual features readily overshadow subtle, sparse force signals, we propose \textbf{ForceFlow}, a force-aware reactive framework for contact-rich manipulation.

To further achieve robust spatial generalization, we integrate ForceFlow with a hierarchical \textbf{Vision-to-Force (V2F) handover mechanism}. The V2F mechanism decouples manipulation into a macro-level \textit{Approach Stage} and a micro-level \textit{Interaction Stage}. During the Approach Stage, a VLM handles target localization and guides the end-effector into the local interaction workspace; once the designated waypoint is reached, V2F executes the handover and ForceFlow takes over as a local expert for closed-loop contact control.

\subsection{Preliminary}
\label{sec:preliminary}

\begin{wrapfigure}{R}{0.5\textwidth}
  \vspace{-1.5cm}
  \centering
  \includegraphics[width=0.9\linewidth]{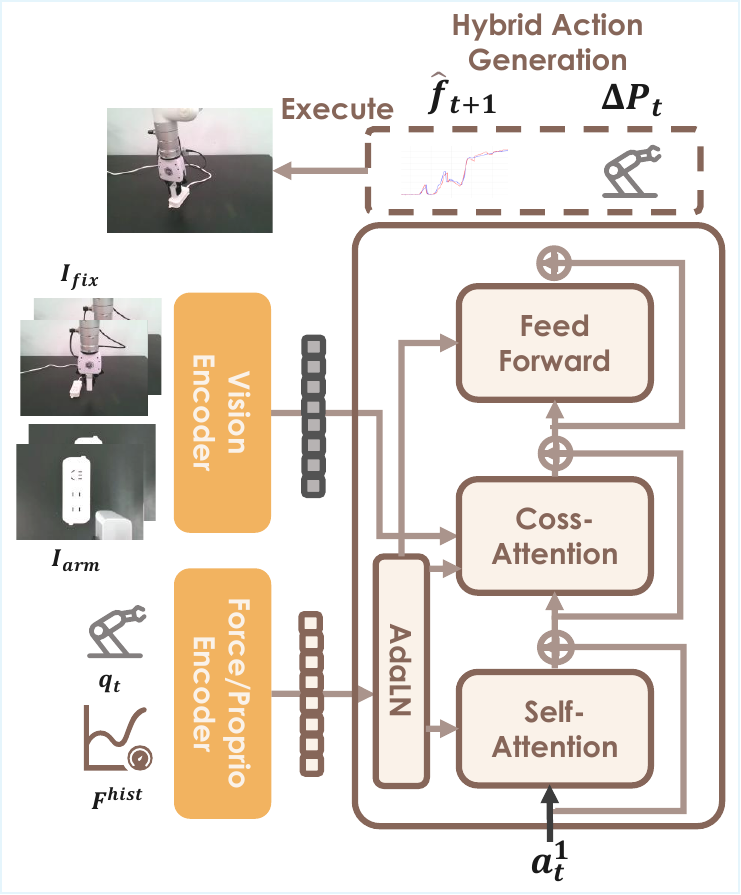} 
  \caption{\textbf{The ForceFlow Architecture.}}
  \label{fig:overview}
  \vspace{-1cm}
\end{wrapfigure}

We consider the task of training a contact-rich policy $\pi_\theta(\mathbf{a}_{t} | \mathcal{O}_t)$ from a set of expert demonstrations $\mathcal{D}$. The multimodal observation at time $t$ is defined as $\mathcal{O}_t = \{ I_{\text{arm},t}, I_{\text{fix},t}, \mathbf{q}_t, \mathbf{F}_{t}^{\text{hist}} \}$, comprising dual-view RGB images, proprioception $\mathbf{q}_t \in \mathbb{R}^{d_q}$, and a force-torque history window $\mathbf{F}_{t}^{\text{hist}} \in \mathbb{R}^{H \times d_f}$. Here, $H$ denotes the temporal history horizon and $d_f$ the dimension of force feedback. Based on $\mathcal{O}_t$, the policy $\pi_\theta$ predicts an action chunk $\mathbf{a}_t$ of length $H^a$.

\subsection{Approach Stage: Pointing Mechanism and V2F Handover}
\label{sec:vlm_pointing}

As discussed in Section 1, contact-rich manipulation requires both spatial and physical generalization, yet these two capabilities demand fundamentally different mechanisms: the former relies on semantic reasoning, while the latter depends on high-frequency force feedback. Coupling both within a single end-to-end policy causes mutual performance degradation. We therefore propose the \textbf{Vision-to-Force (V2F) handover mechanism}, which explicitly divides the manipulation task into a vision-dominant Approach Stage and a force-dominant Interaction Stage, with stage switching triggered by a spatial arrival condition. This design delegates spatial generalization entirely to the VLM's zero-shot reasoning capability, and reserves physical generalization for ForceFlow's force-aware closed-loop regulation.

\textbf{VLM Pointing Mechanism.} During the Approach Stage, we leverage the VLM's open-world spatial reasoning to perform semantic target localization. Given a natural language instruction and a global camera view $I_{\text{fix}}$, the VLM predicts the pixel coordinates $(u, v)$ of the target contact keypoint, isolating the spatial generalization problem as a pure semantic localization task. To equip the VLM with precise localization ability, we construct a visually grounded dataset by manually annotating the 2D coordinates $(u_{\text{gt}}, v_{\text{gt}})$ of target interaction regions on the initial frames of expert demonstrations, formatting them as VQA conversational pairs (image + language instruction $\rightarrow$ coordinates), and fine-tuning the VLM accordingly.

\textbf{V2F Handover and Inference.} At inference time, the VLM predicts the target pixel $(\hat{u}, \hat{v})$, which is deprojected into a 3D approach waypoint in the robot's base frame using depth information and camera intrinsics. A motion planner navigates the end-effector to this region. Upon arrival, the V2F handover is triggered (strictly based on positional criteria), transferring control to the ForceFlow policy for force-aware closed-loop regulation. Since global spatial reasoning has already been completed by the VLM during the Approach Stage, ForceFlow operates solely as a local expert focused on contact dynamics, without the need to model global visual features.

\subsection{Contact Stage: Contact-Driven Flow Matching Policy}
\label{sec:forceflow_policy}

Once the VLM guides the end-effector to the 3D approach waypoint within the local interaction window, the system transitions to the ForceFlow policy for the Interaction Stage. ForceFlow focuses exclusively on high-frequency contact dynamics without bearing the burden of global spatial reasoning. To this end, we design a policy combining an asymmetric multimodal fusion mechanism with a flow matching generative backbone. An overview of the ForceFlow architecture is shown in Figure~\ref{fig:overview}.

\textbf{Asymmetric Multimodal Fusion.}
In contact-rich manipulation, high-dimensional visual features can easily overshadow subtle, low-dimensional force signals during end-to-end training. To prevent this visual dominance, we partition the observation set $\mathcal{O}_t$ into two specialized pathways:
\begin{itemize}[leftmargin=*]
    \item \textbf{Force-Centric Vector Condition ($c_{\text{vec}}$):} To ensure that force signals persistently influence the generation process across all network layers, we encode the temporal force history $\mathbf{F}_{t}^{\text{hist}}$ (of length $H$) together with proprioception $\mathbf{q}_t$ into a unified global representation $c_{\text{vec}}$. This vector is injected into the Diffusion Transformer (DiT) via Adaptive Layer Normalization (AdaLN) \citep{dit}. By modulating feature statistics globally at every network layer, force signals act as a persistent regulatory constraint rather than being marginalized.
    \item \textbf{Visual Sequence Condition ($c_{\text{seq}}$):} Multi-view RGB observations are processed into spatial features. Rather than pooling them into a single vector, we preserve their temporal order to form a sequence condition $c_{\text{seq}}$, integrated via Cross-Attention. This allows the model to selectively attend to relevant spatio-temporal visual cues while remaining consistent with the global force state.
\end{itemize}

\textbf{Hybrid Action Generation via Flow Matching.}
We employ Flow Matching~\citep{flowmatching,cleandiffuser} to learn a deterministic velocity field over a \textbf{hybrid action space} $\mathbf{a}_t = \left[\Delta \mathbf{p}_t, \hat{\mathbf{f}}_{t+1}\right]$. By jointly predicting the motion command $\Delta \mathbf{p}_t$ and the expected contact force $\hat{\mathbf{f}}_{t+1}$, the policy is encouraged to internalize the correlation between robot motion and contact feedback.

We construct a linear probability path interpolating between expert hybrid actions $\mathbf{a}_t^0 \sim p_{\text{data}}(\mathbf{a})$ and a standard Gaussian prior $\mathbf{a}_t^1 \sim \mathcal{N}(0, \mathbf{I})$. For a flow time step $k \in [0, 1]$, the intermediate state is $\mathbf{a}_t^k = (1 - k)\mathbf{a}_t^0 + k \mathbf{a}_t^1$. We parameterize the neural velocity field $v_\theta(\mathbf{a}_t^k, k, c_{\text{vec}}, c_{\text{seq}})$ to approximate the constant target drift $\mathbf{u}_t^k = \mathbf{a}_t^1 - \mathbf{a}_t^0$ using the following objective:
\begin{equation}
    \mathcal{L}_{\text{FM}}(\theta) = \mathbb{E}_{k, \mathbf{a}_t^0, \mathbf{a}_t^1} \left \Vert v_\theta(\mathbf{a}_t^k, k, c_{\text{vec}}, c_{\text{seq}}) - \mathbf{u}_t^k \right\Vert^2
\end{equation}

During inference, we recover the noise-free hybrid action $\mathbf{a}_t^0$ by solving the ODE $d\mathbf{a}_t^k = v_\theta(\mathbf{a}_t^k, k, c_{\text{vec}}, c_{\text{seq}}) dk$ from $k=1$ to $k=0$ using a deterministic numerical solver. At execution time, only the motion command $\Delta \mathbf{p}_t$ is sent to the robot controller; the force prediction $\hat{\mathbf{f}}_{t+1}$ serves as a joint training objective that encourages the network to internalize the coupling between force and motion, rather than directly participating in low-level force control. This ensures physically consistent trajectory generation while enhancing the policy's awareness of contact states through the joint prediction mechanism.

\begin{figure}[t]
  \centering
  \includegraphics[width=\linewidth]{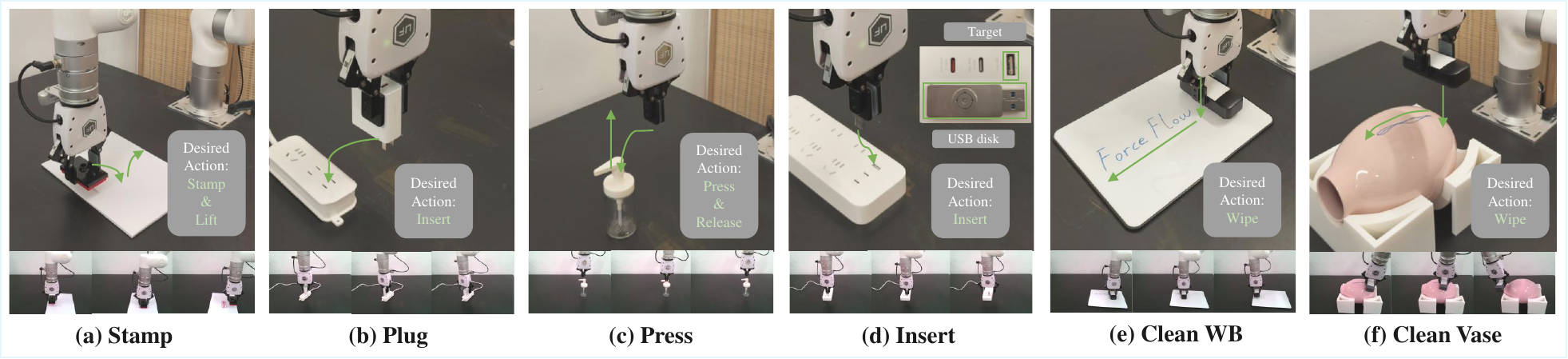}
  % \vspace{-0.1cm}
  \caption{\textbf{Contact-Rich Task Suite.} We evaluate ForceFlow on six diverse manipulation tasks categorized into: (a)-(d) \textbf{Short-Horizon Contact Tasks} requiring precise contact establishment; and (e)-(f) \textbf{Continuous Contact Tasks} demanding consistent normal force tracking along trajectories.}
  \label{fig:task}
\end{figure}

\section{Experiments}
We evaluate ForceFlow on six contact-rich manipulation tasks to answer four key research questions:

\noindent \textbf{RQ1 (Effectiveness):} How does ForceFlow compare against state-of-the-art vision-centric and force-aware baselines (e.g. ForceVLA)?

\noindent \textbf{RQ2 (Force Fidelity):} Can the framework learn to regulate interaction forces and maintain consistency due to the deterministic nature of flow matching?

\noindent \textbf{RQ3 (Generalization):} Does the V2F handover mechanism enhance robustness in OOD scenarios?

\noindent \textbf{RQ4 (Ablation):} What are the individual contributions of force history and active force prediction?

Additionally, we qualitatively demonstrate ForceFlow's capability in continuous force regulation through a cucumber peeling task.

\subsection{Experimental Setup}

\textbf{Task Suite (\Cref{fig:task}).}
We select six tasks categorized into: (1) short-horizon contact tasks that require precise modulation to establish contact (Stamping, Plug/USB Insertion, Press Button); and (2) continuous contact tasks which demand consistent normal force (Clean Whiteboard, Clean Vase). Detailed task settings are provided in Appendix~\ref{app:task}.

We employ two complementary metrics:

\noindent \textbf{Success Rate (SR):} The percentage of successful trials ($N=20$) based on task-specific completion criteria (e.g., successful insertion or clean wiping).

\noindent \textbf{Force Fidelity (MAE Cost):} To quantify physical compliance and stability, we measure the deviation between the policy's interaction force and the expert's reference. We define the metric $\mathcal{J}_{\text{force}}$ as:
\begin{equation}
    \mathcal{J}_{\text{force}} = \frac{1}{N} \sum_{i=1}^{N} \left\vert \hat{F}_{\text{policy}}^{(i)} - F_{\text{expert}} \right\vert
\end{equation}
where $N=20$ is the number of trials. The force statistic $\hat{F}$ is task-dependent: for \textbf{Short-Horizon Contact Tasks}, it denotes the \textit{peak contact force} ($\max_t \|\mathbf{f}_t\|$) to capture impact intensity; for \textbf{Continuous Contact Tasks}, it represents the \textit{average effective force} calculated only during the contact phase (where $\|\mathbf{f}_t\| > 5\text{N}$) to evaluate tracking consistency.

% \paragraph{Overall Comparison of Force Cost} % ???????????????????
\begin{table}[t]
    \tiny
    \centering
    \caption{\textbf{Quantitative Comparison of Task Success Rates (SR).} We report the success rate (\%) over 20 trials for each task. \textbf{Bold} indicates the best performance.}
    \label{tab:success_rate}
    \vspace{-0pt}
    \resizebox{0.93\textwidth}{!}{%
    \begin{tabular}{lccccccc}
        \toprule
        \textbf{Method} & \textbf{Stamp} & \textbf{Plug} & \textbf{Press} & \textbf{Insert} & \textbf{Clean WB} & \textbf{Clean Vase} & \textbf{Avg.} \\
        \midrule
        $\pi_{0.5}$  & 0\% & 60\% & 30\% & 45\% & 10\% & 0\% & 24.17\% \\
        ACT & 0\% & 30\% & 5\% & 0\% & 15\% & 0\% & 8.33\% \\
        Diffusion Policy & 0\% & 40\% & 20\% & 50\% & 75\% & 0\% & 30.83\% \\
        ForceVLA & 20\% & 70\% & 65\% & 15\% & 100\% & 0\% & 45\% \\
        \midrule
        ForceFlow (w/o Force) & 20\% & 75\% & 0\% & 40\% & 100\% & 30\% & 44.17\% \\
        \textbf{ForceFlow (Ours)} & \textbf{85\%} & \textbf{90\%} & \textbf{90\%} & \textbf{60\%} & \textbf{100\%} & \textbf{65\%} & \textbf{81.67\%} \\
        \bottomrule
    \end{tabular}%
    }
\end{table}

\begin{table*}[t]
    \tiny
    \centering
    \caption{\textbf{Force Fidelity Analysis (MAE Cost).} The table reports the MAE between the executed force and the expert demonstration force. The unit is Newton (N). \textbf{Bold} indicates the lowest cost (highest fidelity). \textbf{Lower is better.} }
    \label{tab:force_cost}
    \vspace{-0.1cm}
    \resizebox{0.93\textwidth}{!}{%
    \begin{tabular}{lccccccc}
        \toprule
        \textbf{Method} & \textbf{Stamp} & \textbf{Plug} & \textbf{Press} & \textbf{Insert} & \textbf{Clean WB} & \textbf{Clean Vase} & \textbf{Avg.} \\
        \midrule
        $\pi_{0.5}$ & 31.99 & 21.41 & 17.39 & 50.89 & 11.93 & 7.87 & 23.58 \\
        ACT & 31.86 & 25.54 & 31.81 & 38.71 & 11.91 & 30.45 & 28.38 \\
        Diffusion Policy & 32.26 & 15.79 & 24.86 & 23.85 & 8.22 & 23.56 & 21.42 \\
        ForceVLA & 30.03 & 9.59 & 30.94 & 37.82 & 20.16 & 11.29 & 23.31 \\
        \midrule
        ForceFlow (w/o Force) & 30.03 & 13.36 & 37.50 & 34.75 & 7.16 & 13.24 & 22.67 \\
        \textbf{ForceFlow (Ours)} & \textbf{10.61} & \textbf{3.58} & \textbf{5.03} & \textbf{21.79} & \textbf{4.59} & \textbf{3.76} & \textbf{8.23} \\
        \bottomrule
    \end{tabular}%
    }
    \vspace{-0.5cm}
\end{table*}

\subsection{RQ1: Effectiveness in Contact-Rich Tasks}

We compare ForceFlow against $\pi_{0.5}$~\citep{pi0.5}, ACT~\citep{act}, Diffusion Policy~\citep{diffusion_policy}, ForceVLA~\citep{forcevla}, and the ablation baseline ForceFlow (w/o Force). As presented in Table~\ref{tab:success_rate}, ForceFlow achieves an average success rate of \textbf{81.67\%}, significantly outperforming the best baseline (ForceVLA, 45\%). We analyze this performance gain through three functional paradigms of force perception:

\noindent \textbf{Resolving Visual Ambiguity (Force as Primary Modality).} In tasks like \textit{Stamping} and \textit{Press Button}, vision-centric baselines collapse (0--30\% SR) because they cannot perceive environmental properties such as paper stack thickness (ranging from 1 to 50 sheets) or varying spring constants. They tend to converge to a mean terminal height, leading to either insufficient pressure or excessive collision. In contrast, ForceFlow exploits the 10-step force history to detect the exact onset of resistance. By treating force-torque profiles as the ground truth for state transitions, it achieves high success rates (85\% and 90\%) regardless of visual uncertainties in object height or stiffness.

\noindent \textbf{Navigating Geometric Constraints (Force as Auxiliary Modality).} For \textit{Plug} and \textit{USB Insertion} involving sub-millimeter tolerances, visual feedback often erroneously indicates completion when the connector is actually stalled by friction. ForceFlow achieves 60\% SR on the challenging USB task (where baselines largely fail) by perceiving the reactive torque $\tau$ generated by misalignment. Combined with active force prediction, the policy generates subtle sliding and wiggling motions to feel the opening and dynamically align the components, overcoming geometric jamming that vision-only policies cannot resolve.

\noindent \textbf{Stability in Continuous Interaction (Force as Regulatory Modality).} The advantage of ForceFlow is most pronounced in tracking non-linear geometries. While ForceVLA performs well on the planar \textit{Whiteboard} (100\%), it fails completely on the curved \textit{Clean Vase} (0\%) due to the continuously changing surface normals. ForceFlow utilizes active force prediction as a regulatory mechanism to proactively compliance-match the surface curvature. This allows it to maintain a stable interaction envelope on irregular 3D surfaces, achieving a 65\% success rate.

\subsection{RQ2: Force Regulation and Fidelity (Cost Analysis)}

To answer RQ2, we conduct an in-depth analysis of the model's force control behavior from three perspectives: statistical force deviation (Force Cost), stability across trials, and the consistency between predicted and real contact forces during single execution. ForceFlow significantly reduces Force Cost across all tasks. The average Force Cost drops from the 20--30 N range of vision-dominant models to 8.23 N. The reduction is particularly notable ($>50\%$) in instantaneous or semi-instantaneous contact tasks such as Stamp, Plug, and Press. Although ForceFlow (w/o Force) can still complete operations in some tasks, its Force Cost is significantly higher than the full model, highlighting that Success Rate alone is an incomplete proxy for force quality, as geometrically correct actions can still deviate severely from expert force distributions.

\begin{figure}[H]
  \vspace{-0.2cm}
  \centering
  
  % 左侧第一个独立图
  \begin{minipage}[b]{0.48\textwidth}
    \centering
    \includegraphics[width=\linewidth]{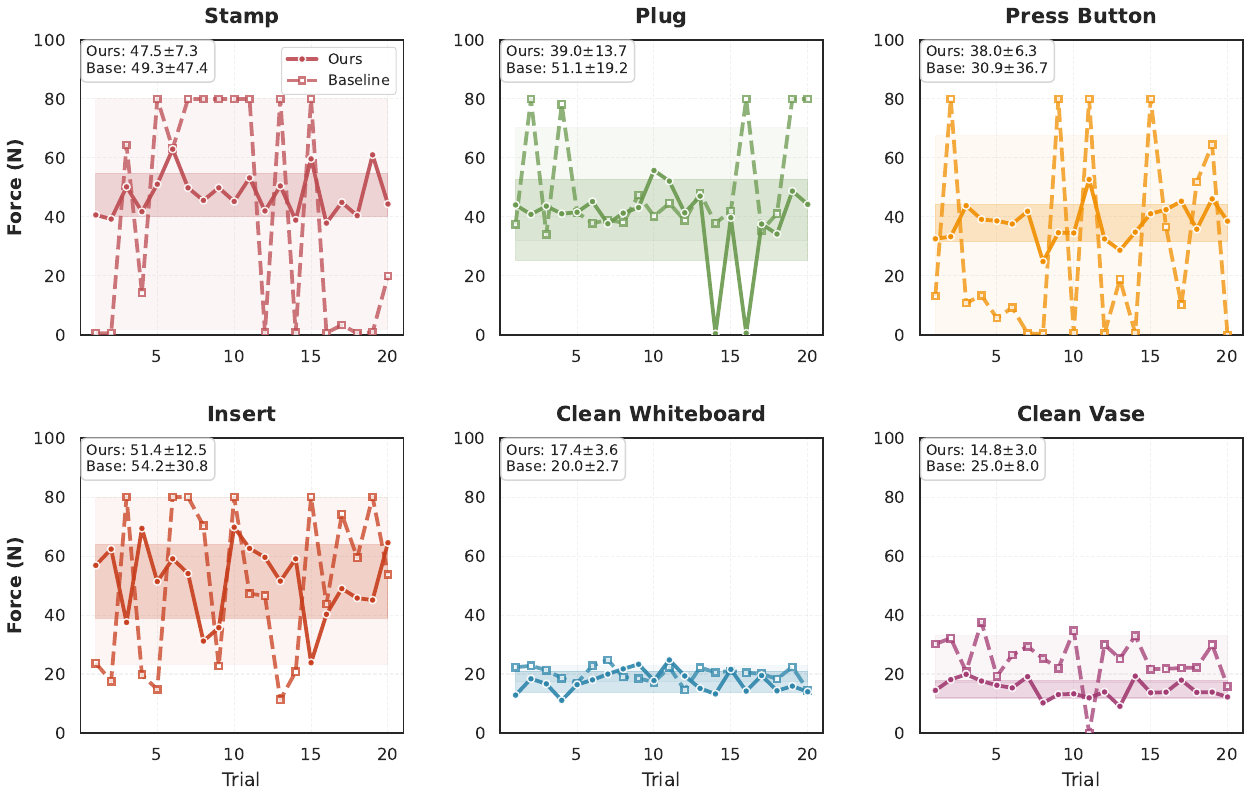}
    \vspace{-0.5cm}
    \caption{\textbf{Statistical Force Distribution}. Maximum or average interaction forces for six representative tasks across 20 independent trials. Error bars/shading indicate standard deviation, highlighting the stability of the learned force regulation strategy.}
    \label{fig:compare}
  \end{minipage}
  \hfill % 在两个 minipage 之间自动填充弹性空白，把它们向两边推开
  % 右侧第二个独立图
  \begin{minipage}[b]{0.48\textwidth}
    \centering
    \includegraphics[width=\linewidth]{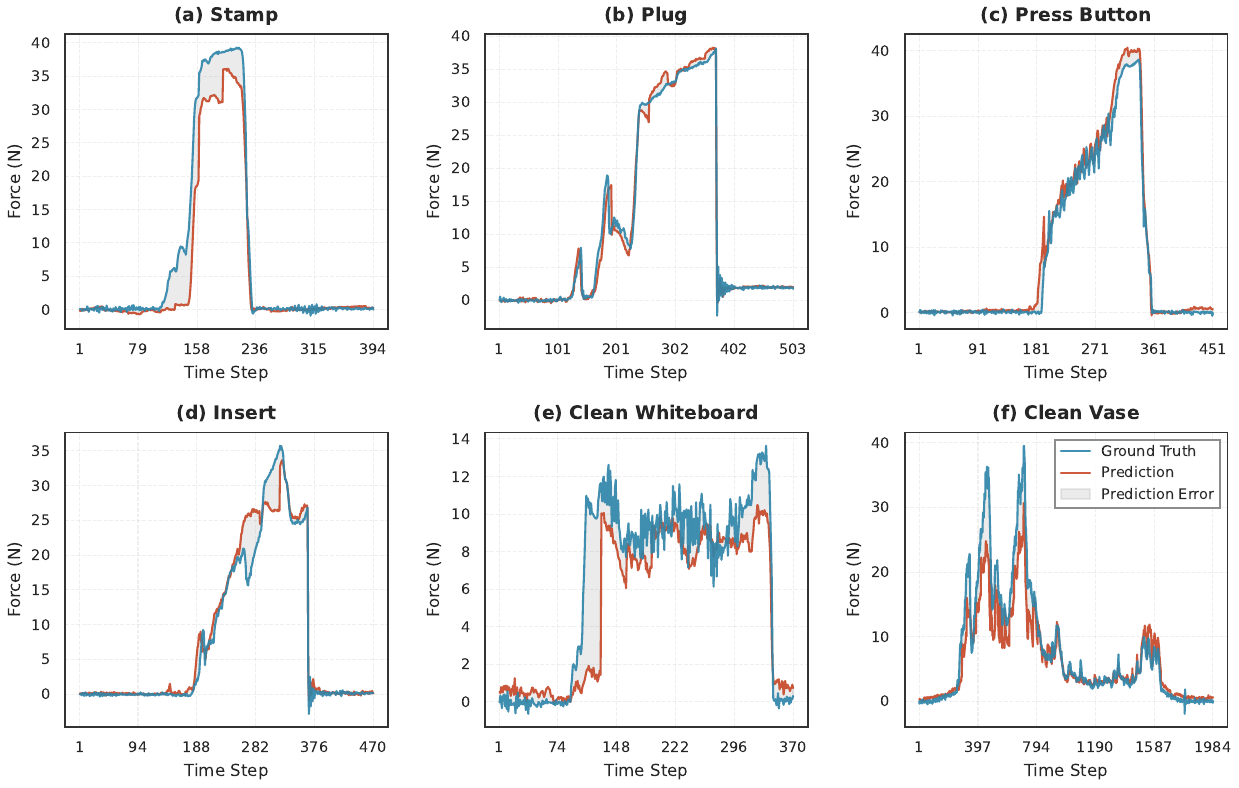}
    % \vspace{-0.2cm}
    \caption{Comparison of Predicted and Measured Forces. Alignment between the predicted (red) and measured ground truth (blue) contact forces. Shaded areas represent prediction error, demonstrating high temporal fidelity during interaction.}
    \label{fig:predict}
  \end{minipage}
  
  \vspace{-0.3cm}
\end{figure}

\paragraph{Stability Across Trials (20 Test Statistics)}
Figure~\ref{fig:compare} illustrates the Maximum Contact Force and Average Effective Force for ForceFlow and ForceFlow (w/o Force) across 20 independent trials. ForceFlow's force output is highly consistent across trials. Fluctuations are minimal, indicating that the model has learned a \textit{task-dependent force regulation strategy} rather than memorizing a single trajectory.
Conversely, the model without force feedback exhibits obvious instability. ForceFlow (w/o Force) shows distinct issues: ForceFlow (w/o Force) exhibits severe instability (e.g., violent oscillations and abnormal spikes) in contact establishment, stably excessive forces in continuous tracking, and clear open-loop characteristics (e.g., insensitivity to paper thickness in Stamping).
This demonstrates that relying solely on vision and proprioception prevents the model from forming a stable closed-loop mechanism for contact force regulation.

\paragraph{Consistency of Force Prediction in Single Execution}
Figure~\ref{fig:predict} displays the alignment between ForceFlow's predicted contact force curve (red) and the real measured force (blue) across six representative tasks.
In short-contact tasks, predicted forces track real forces synchronously during both contact establishment and release. This suggests that the model can infer the impending contact state via force history, vision, and proprioception.
In continuous tasks (Clean Whiteboard/Clean Vase), the predicted force smoothly adjusts following surface changes, without high-frequency jitter, proving the model learns state-based continuous force regulation.

\subsection{RQ3: Generalization to OOD Environments}

This section evaluates the robustness of ForceFlow across complex real-world variations by decoupling visual localization from force regulation. We analyze two critical dimensions of generalization: (1) \textbf{Physical Interaction Generalization}, where contact objects or tools change but spatial locations remain fixed, and (2) \textbf{Spatial Generalization}, where target objects are placed in OOD spatial locations.

\begin{figure}[htbp]
\vspace{-0.1cm}
  \centering
  \includegraphics[width=0.95\linewidth]{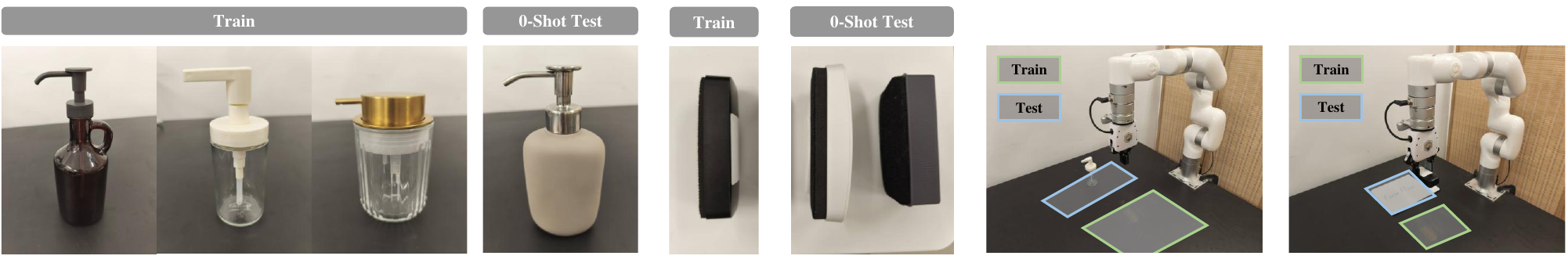}
  \vspace{-0.3cm}
  \caption{\textbf{OOD Evaluation Setup.} (a) \textbf{Physical Interaction Generalization:} Unseen objects (e.g., distinct bottles, different erasers)  used for zero-shot testing. (b) \textbf{Spatial Generalization:} The testing workspace (marked ``Test'') is spatially disjoint from the training distribution (marked ``Train''), requiring V2F for localization.}
  \label{fig:ood}
  \vspace{-0.2cm}
\end{figure}

\paragraph{Physical Interaction Generalization: Zero-Shot Adaptability.}
We evaluate physical adaptability by replacing training tools with unseen variants without fine-tuning (see Figure~\ref{fig:ood}a). We conducted 10 independent trials per task using objects with distinct physical properties: for \textbf{Clean Vase} and \textbf{Whiteboard}, we introduced two novel erasers with significantly different stiffness and thickness; for \textbf{Press Button}, we utilized a new bottle with a distinct height and spring trigger threshold.

As shown in Table~\ref{tab:force_ood}, vision-centric baselines fail entirely when tool friction or object elasticity changes, as they cannot perceive high-frequency physical feedback. In contrast, ForceFlow maintains high success rates by adaptively adjusting contact forces through its temporal force perception and active prediction mechanisms. By encoding $F_{hist}$, the model perceives contact states and achieves proactive compliance even when the physical interaction parameters deviate from the training distribution.

\paragraph{Spatial Generalization: Hierarchical Grounding.}
To evaluate robustness against spatial shifts, we configure a testing workspace that is spatially disjoint from the training distribution, as illustrated in Figure~\ref{fig:ood}(b). To tackle large spatial variations, we implement V2F, specifically utilizing Embodied-R1~\cite{er1} as the high-level planner. This hierarchical strategy allows the system to localize targets in OOD positions significantly outside the training distribution.

Results in Table~\ref{tab:vision_ood} highlight the necessity of V2F integration. Without semantic guidance, low-level policies (including standalone baselines) fail to locate objects in OOD regions. By utilizing V2F, the system predicts semantic pixel coordinates from the global view $I_{fix}$ to guide initial movement. Once the end-effector enters the local interaction window, ForceFlow takes over for high-frequency regulation. This separation of concerns ensures that the model handles diverse backgrounds and spatial shifts while maintaining the determinism required for contact-intensive tasks.

\begin{table}[h]
    \centering
    
    % --- 第一行：Caption ---
    \begin{minipage}[b]{0.33\linewidth}
        \caption{\textbf{Physical Interaction Generalization (SR \%).}}
        \label{tab:force_ood}
    \end{minipage}%
    \hfill
    \begin{minipage}[b]{0.29\linewidth}
        \caption{\textbf{Spatial Generalization (SR \%).}}
        \label{tab:vision_ood}
    \end{minipage}%
    \hfill
    \begin{minipage}[b]{0.34\linewidth} 
        \caption{\textbf{Ablation Study on the Stamp Task.}}
        \label{tab:ablation_stamp}
    \end{minipage}
    
    \vspace{0.15cm} 
    
    % 【关键改动 1】：把 arraystretch 放在外面，统一全局拉伸比例，保证表头和表尾高度绝对一致！
    \renewcommand{\arraystretch}{1.15}
    
    % --- 第二行：表格主体 ---
    \begin{minipage}[t]{0.33\linewidth}
        \vspace{-0.8cm}
        \centering
        \tiny 
        \setlength{\tabcolsep}{1.5pt} 
        \begin{tabular*}{\linewidth}[t]{@{\extracolsep{\fill}} lccc @{}}
            \toprule
            \textbf{Method} & \textbf{Press} & \textbf{Clean WB} & \textbf{Clean Vase} \\
            \midrule
            % 【关键改动 2】：表1缺1行高度，通过 \\[0.3em] 均匀平摊给这4行
            $\pi_{0.5}$ & 0\% & 0\% & 0\% \\[0.3em]
            ACT & 0\% & 0\% & 0\% \\[0.3em]
            Diffusion Policy & 0\% & 0\% & 0\% \\[0.3em]
            ForceVLA & 40\% & 90\% & 0\% \\[0.3em]
            \midrule
            \textbf{ForceFlow(Ours)} & \textbf{80\%} & \textbf{100\%} & \textbf{60\%} \\
            \bottomrule
        \end{tabular*}
    \end{minipage}%
    \hfill
    \begin{minipage}[t]{0.29\linewidth}
        \vspace{-0.8cm}
        \centering
        \tiny 
        \setlength{\tabcolsep}{1.5pt} 
        \begin{tabular*}{\linewidth}[t]{@{\extracolsep{\fill}} lccc @{}}
            \toprule
            \textbf{Method} & \textbf{Press} & \textbf{Plug} & \textbf{Clean WB} \\
            \midrule
            % 表2是基准（5行），正常换行即可
            $\pi_{0.5}$ & 0\% & 0\% & 0\% \\
            ACT & 0\% & 0\% & 0\% \\
            Diffusion Policy & 0\% & 0\% & 0\% \\
            ForceVLA & 0\% & 0\% & 0\% \\
            ForceFlow & 0\% & 0\% & 0\% \\
            \midrule
            \textbf{ForceFlow + V2F} & \textbf{40\%} & \textbf{10\%} & \textbf{50\%} \\
            \bottomrule
        \end{tabular*}
    \end{minipage}%
    \hfill
    \begin{minipage}[t]{0.34\linewidth}
        \vspace{-0.8cm}
        \centering
        \tiny 
        \setlength{\tabcolsep}{1.5pt} 
        \begin{tabular*}{\linewidth}[t]{@{\extracolsep{\fill}} lcc @{}}
            \toprule
            \textbf{Variant} & \textbf{SR (\%)} & \textbf{Cost (N)} \\
            \midrule
            % 【关键改动 3】：表3缺2行高度，通过 \\[0.8em] 均匀平摊给这3行
            w/o Force History(1-step) & 55\% & 15.50 \\[0.8em]
            w/o Force Prediction & 80\% & 12.52 \\[0.8em]
            w/o Both(1-step + No Pred) & 40\% & 18.21 \\[0.8em]
            \midrule
            \textbf{ForceFlow(Full)} & \textbf{85\%} & \textbf{10.61} \\
            \bottomrule
        \end{tabular*}
    \end{minipage}
    
\end{table}

\subsection{RQ4: Ablation Study}

This section investigates the individual contributions of temporal force history and active force prediction to the overall performance and stability of ForceFlow. We evaluate three ablation variants across our task suite to quantify the impact of our force-centric design choices.

\begin{figure}[t]
  \centering
  \includegraphics[width=0.95\textwidth]{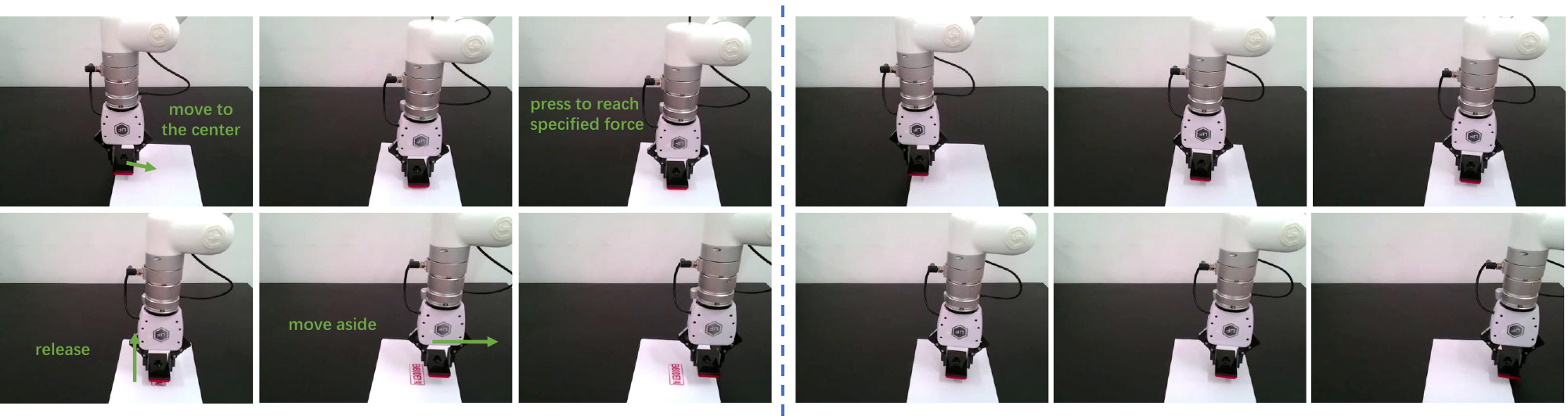} 
  \caption{Qualitative comparison of force regulation in the Stamping task. The upper part illustrates the stable execution of ForceFlow (Full), while the lower part shows the w/o Force History (1-step) variant. Without temporal context, the single-step policy suffers from severe oscillations and force spikes due to sensor noise, whereas ForceFlow maintains a stable and smooth contact profile.}
  \label{fig:ablation_vis}
  \vspace{-10pt}
\end{figure}

As detailed in Table~\ref{tab:ablation_stamp}, the results reveal different roles for each component. Temporal force history proves decisive for task success. Replacing the 10-step window with a single reading causes a drastic SR drop (85\% to 55\%). This performance degradation is due to the fact that instantaneous force measurement is inherently susceptible to interference from contact instability or noise from the sensor itself. This instability is visually corroborated in Figure~\ref{fig:ablation_vis}. The 1-step variant exhibits significant high-frequency jitter and abrupt force spikes during contact, indicating an inability to distinguish between signal noise and true physical feedback. In contrast, ForceFlow leverages the historical window to smooth out these fluctuations, resulting in a stable interaction. Without temporal context to filter these fluctuations, the policy fails to reliably estimate the true interaction state. In contrast, active force prediction acts as a compliance regularizer. Removing the prediction head results in a marginal SR decline (80\%) but a notable increase in Force Cost (10.61N to 12.52N), suggesting that it smooths the interaction rather than dictating feasibility. The significant degradation when both are removed (40\% SR) confirms their synergistic necessity: history ensures correct decision-making, while prediction refines execution quality.

\begin{figure}[H]
  \centering
  \includegraphics[width=\linewidth]{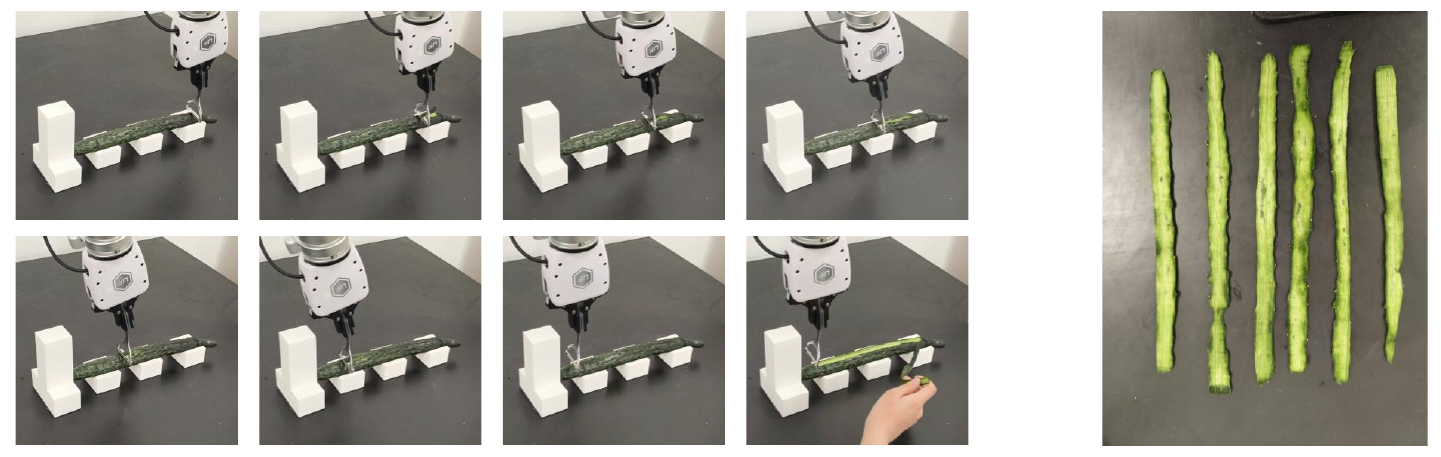}
  \caption{\textbf{Qualitative results of cucumber peeling.} (Left) ForceFlow proactively regulates force to track the contour. (Right) Uniform, unbroken peeled strips evidence a consistent cutting depth.}
  \label{fig:peel}
\end{figure}

\subsection{Qualitative Analysis: Continuous Force Regulation in Cucumber Peeling}

We qualitatively evaluate ForceFlow on a cucumber peeling task (Fig.~\ref{fig:peel}, Left), which requires maintaining precise normal force along a varying-stiffness surface stabilized on V-groove supports.

Visual ambiguity often causes vision-centric baselines to miss contact or apply excessive force, jamming the tool. Conversely, ForceFlow utilizes temporal force history to accurately detect the onset of resistance upon initial contact, establishing optimal cutting depth without over-pressing.

During sliding, the model leverages active force prediction to dynamically adjust downward pressure, seamlessly adapting to geometric variations (e.g., local bumps or tapering ends) to track the contour. The resulting uniform, unbroken peeled strips (Fig.~\ref{fig:peel}, Right) evidence ForceFlow's ability to maintain a stable interaction envelope and consistent cutting depth.

\section{Conclusion}
In this paper, we propose ForceFlow, a force-aware reactive framework based on the Flow Matching method for contact-rich manipulation tasks. Through the temporal force history and a joint prediction mechanism, ForceFlow enables robots to achieve deep synergy between visual perception and physical contact dynamics. The Vision-to-Force (V2F) mechanism effectively decouples fine manipulation into a VLM-guided approach stage and a force-dominant interaction stage, ensuring the system exhibits excellent robustness against spatial and physical distribution shifts. Experimental results on six real-world tasks demonstrate that ForceFlow improves the success rate by 37\% compared with state-of-the-art baselines, while demonstrating exceptional zero-shot OOD generalization capability. Despite these promising results, ForceFlow still has certain limitations. For instance, the current framework relies on high-fidelity force/torque sensors, which may restrict its deployment on low-cost robotic platforms. Furthermore, future research can develop an adaptive V2F switching framework to further enhance the overall manipulative dexterity of embodied robots in unstructured environments.

% ============================================================================
% References
% ============================================================================
\bibliography{references}

% ============================================================================
% Appendix (uncomment if needed)
% ============================================================================
% \appendix
% \section{Additional Results}
\newpage
\appendix
\onecolumn

\section{Detailed Experimental Setup}
\label{app:task}

% Detailed Training Hyperparameters Table for ICML Paper Appendix
% Two-column version with additional architectural details

\begin{table*}[h]
\centering
\caption{Detailed Training Hyperparameters and Model Architecture}
\label{tab:hyperparameters_detailed}
\small
\begin{tabular}{llll}
\toprule
\textbf{Category} & \textbf{Hyperparameter} & \textbf{Value} & \textbf{Description} \\
\midrule
\textit{Training} & Batch Size & 64 & Per-device batch size \\
& Max Steps & 100,000 & Total training steps \\
& Precision & bf16-mixed & Mixed precision training \\
& Gradient Accumulation & 1 & Accumulation batches \\
& Checkpoint Interval & 5,000 & Steps between checkpoints \\
& Random Seed & 0 & Reproducibility seed \\
\midrule
\textit{Data} & Image Resolution & $320 \times 240$ & RGB image size \\
& Observation Horizon ($H_o$) & 2 & Number of observation frames \\
& Action Horizon ($H_a$) & 64 & Action sequence length \\
& Force History ($H_{\text{force}}$) & 10 & Historical force steps \\
& State Dimension & 7 & 6D pose + 1D gripper \\
& Force Dimension & 6 & 3D force + 3D torque \\
& Action Dimension & 13 & 6D pose + gripper + 6D force \\
& Data Workers & 8 & Parallel data loading threads \\
\midrule
\textit{DiT-1D} & Model Dimension & 384 & Base feature dimension \\
& Attention Heads & 6 & Multi-head attention \\
& Transformer Depth & 12 & Number of transformer blocks \\
& Vector Embedding & 256 & Low-dim + force embedding \\
& Sequence Embedding & 512 & Dual-view image embedding \\
& Head Type & MLP & Output head architecture \\
& Cross-Attention & Yes & Visual-to-action attention \\
& AdaLN & Yes & Adaptive Layer Normalization \\
& Timestep Embedding & Fourier & Scale: 0.2 (untrainable) \\
& Activation & SiLU & Swish activation function \\
\midrule
\textit{Condition Encoder} & Visual Backbone & ResNet-18 & Pre-trained, dual views \\
& Image Embedding & $2 \times 256$ & Arm view + fixed view \\
& Low-dim Encoder & 2-layer MLP & Hidden: 64 $\rightarrow$ 64 \\
& Force Encoder & 2-layer MLP & $(H_{\text{force}} \times 6) \rightarrow 128$ \\
& Total Vector Dim & 256 & $64 \times H_o + 128$ \\
& Dropout & 0.0 & No dropout in encoder \\
\midrule
\textit{Diffusion} & Algorithm & Flow Matching & Continuous-time diffusion \\
& Sampling Steps & Variable & Inference-time adjustable \\
& Normalization & MinMax & State and action normalization \\
& Image Normalization & $[-1, 1]$ & Mean: 0.5, Std: 0.5 \\
\bottomrule
\end{tabular}
\end{table*}

% Note: Use \usepackage{booktabs} in your preamble

\subsection{Task Description and Challenges}
We evaluate our framework on six real-world contact-rich manipulation tasks, categorized into short-horizon contact establishment and continuous contact maintenance.

\noindent \textbf{Short-Horizon Contact Tasks.} These tasks require establishing contact with appropriate target forces; insufficient or excessive force leads to failure.
\begin{adjustwidth}{1em}{0em}
\textbf{Stamping (Visual Ambiguity):} The robot stamps stacks of paper of varying thicknesses (1 sheet to $\sim$5 cm). Since stack height is visually indistinguishable, the policy must rely on real-time force feedback to regulate vertical motion and apply optimal stamping pressure.
\end{adjustwidth}

\begin{adjustwidth}{1em}{0em}
\textbf{Plug Insertion (Spatial Randomness):} The robot inserts a power plug into a socket. Initial poses are randomized, and the socket is subject to random deflections. Success requires coarse visual alignment followed by force-guided insertion to overcome contact friction.
\end{adjustwidth}

\begin{adjustwidth}{1em}{0em}
\textbf{USB Insertion (Tight Tolerance):} Similar to the plug task but with sub-millimeter geometric tolerances. The policy must utilize subtle torque cues to ``feel'' the narrow opening of the USB-A port and complete insertion without jamming.
\end{adjustwidth}

\begin{adjustwidth}{1em}{0em}
\textbf{Press Button (Cross-Object):} The robot interacts with three distinct spring-loaded sanitizer bottles. Due to significant variations in height and design, the policy must adaptively regulate contact force to complete a full press-and-release cycle.
\end{adjustwidth}

\noindent \textbf{Continuous Contact Tasks.} These tasks require maintaining stable interaction forces over a longer time horizon.
\begin{adjustwidth}{1em}{0em}
\textbf{Clean Whiteboard (Planar Constant Force):} The robot wipes markings off a fixed whiteboard surface. This requires maintaining a continuous normal force along a 2D planar trajectory to ensure effective cleaning.
\end{adjustwidth}

\begin{adjustwidth}{1em}{0em}
\textbf{Clean Vase (Geometric Complexity):} The robot cleans marks off a curved 3D surface. Although the mount is fixed, the vase may be subject to perturbations, leading to variations in contact normals. This task evaluates force regulation against non-linear, unknown geometries.
\end{adjustwidth}

\subsection{Data Collection and Training Setup}
\begin{figure}[h]
  \centering
  \includegraphics[width=0.5\linewidth]{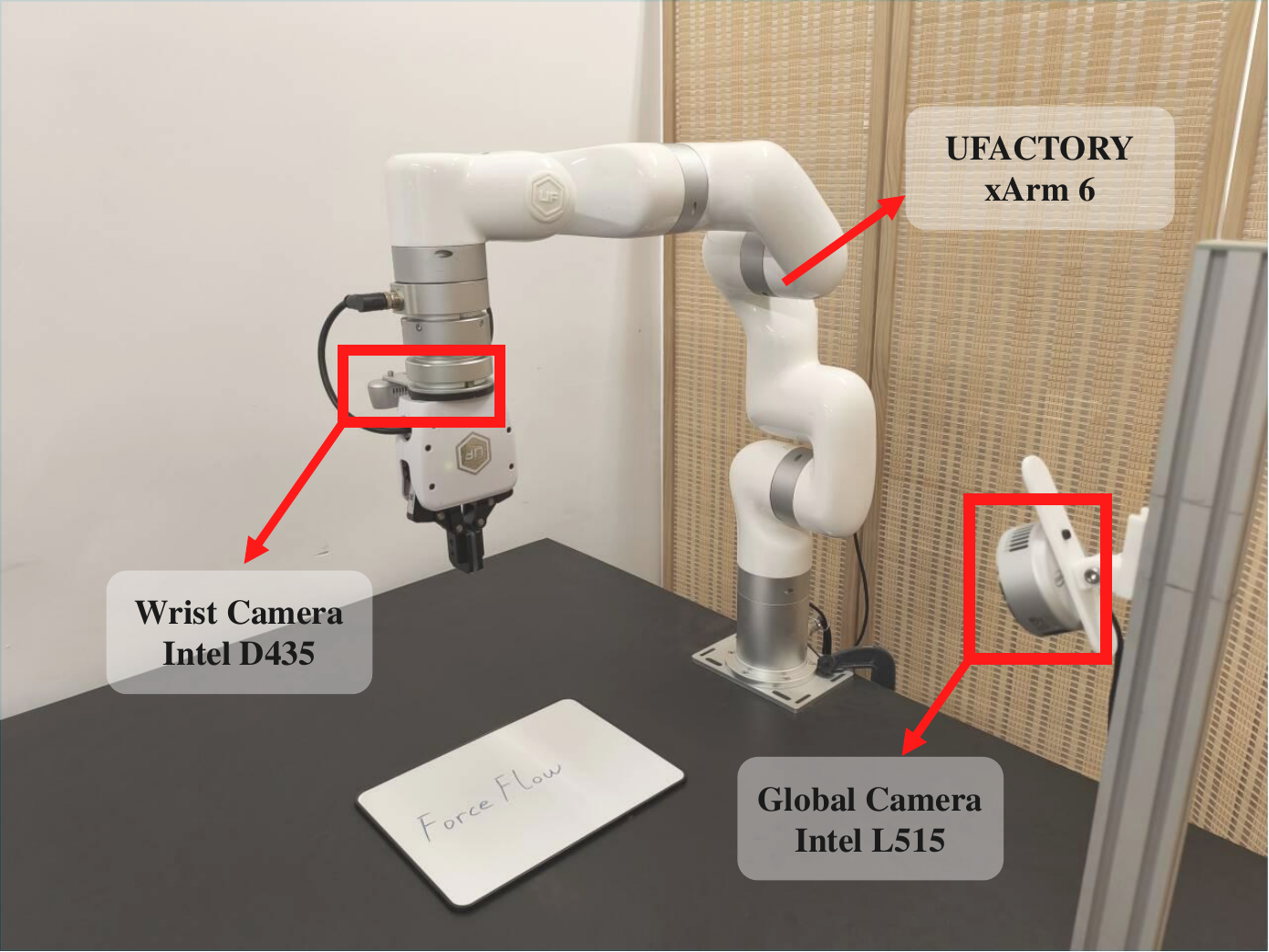} 
  \caption{Hardware System}
  \label{fig:hardware_setup}
\end{figure}

The hardware setup consists of a 6-Dof UFactory xArm6 robotic arm with a 1-Dof gripper, an Intel RealSense L515 camera for the global view, and a D435 camera for the wrist view. Expert demonstrations were collected using two methods: a 3Dconnexion SpaceMouse and a Meta Quest Pro VR headset. The hardware platform is illustrated in Fig. \ref{fig:hardware_setup}.

For each task, we collected 50--100 expert demonstrations via teleoperation at a control frequency of 30 Hz. The dataset includes synchronized dual-view $320 \times 240$ RGB observations, 7D proprioceptive states (6D pose and 1D gripper), and 10-step force-torque histories. Our policy is trained using a high-capacity Diffusion Transformer (DiT) backbone to regress the continuous rectified velocity field.

The models were trained on a compute node equipped with 4$\times$ NVIDIA GeForce RTX 4090 GPUs (24~GB VRAM each), 48 physical CPU cores, and 283~GB system RAM. We employed the AdamW optimizer ($\beta_1 = 0.9, \beta_2 = 0.999$, weight decay = 0.01) with a cosine learning rate schedule starting from $1 \times 10^{-4}$. Training utilized a batch size of 64 with a gradient accumulation of 1 (effective batch size 64), completing 100,000 steps in approximately 8--10 hours per task. To ensure training stability, we employed bfloat16 mixed precision with gradient clipping ($\|\nabla\| = 1.0$). The model architecture and key hyperparameters are detailed in Table~\ref{tab:hyperparameters_detailed}.

\subsection{Inference and Evaluation Protocol}
During inference, actions are generated by deterministically integrating the learned velocity field using an Ordinary Differential Equation (ODE) solver. The policy predicts a 64-step action chunk, executing the first 32 steps before replanning to ensure temporal smoothness. Each task is evaluated over 20 independent trials with randomized initial conditions. Task success is determined by task-specific completion criteria (e.g., full insertion or a clear imprint), while force fidelity is assessed by measuring deviations from expert force profiles.

\section{Mathematical Formulation}
\label{app:math_flow}

This section provides the formal mathematical grounding for the \textbf{ForceFlow} framework, detailing the construction of the probability path, the asymmetric conditioning mechanism, and the joint optimization for active compliance.

\subsection{Linear Probability Path and Velocity Field}
To ensure the trajectory smoothness and determinism required for stable physical contact, ForceFlow employs Continuous Rectified Flow (CRF). 
Consistent with Section~\ref{sec:preliminary}, we define the hybrid action space at task time step $t$ as $\mathbf{a}_t = [\Delta \mathbf{p}_t, \hat{\mathbf{f}}_{t+1}] \in \mathcal{A}$, where $\Delta \mathbf{p}_t \in \mathbb{R}^{d_p}$ represents the motion command (pose increments and gripper status) and $\hat{\mathbf{f}}_{t+1} \in \mathbb{R}^{d_f}$ represents the target contact force for active compliance.

We construct a linear probability path between the expert action distribution $\mathbf{a}_t^0 \sim p_{\text{data}}$ and a standard Gaussian prior $\mathbf{a}_t^1 \sim \mathcal{N}(0, \mathbf{I})$. For a flow time step $k \in [0, 1]$ (superscript), the intermediate state $\mathbf{a}_t^k$ is defined as:
\begin{equation}
    \mathbf{a}_t^k = (1 - k)\mathbf{a}_t^0 + k \mathbf{a}_t^1
\end{equation}
This interpolation defines a rectified flow with a constant ground-truth velocity field $\mathbf{u}_t^k$:
\begin{equation}
    \mathbf{u}_t^k(\mathbf{a}_t^k, k) = \frac{d\mathbf{a}_t^k}{dk} = \mathbf{a}_t^1 - \mathbf{a}_t^0
\end{equation}

\subsection{Asymmetric Multimodal Fusion Architecture}
To prevent high-dimensional visual features from drowning out sparse force signals, we implement an asymmetric fusion strategy within the DiT blocks. The conditioning comprises a vector condition $c_{\text{vec}}$ (derived from proprioception $\mathbf{q}_t$ and force history $\mathbf{F}_{t}^{\text{hist}}$) and a sequence condition $c_{\text{seq}}$ (derived from multi-view visual observations).

\paragraph{Global Regulation via AdaLN.} The vector condition $c_{\text{vec}}$ modulates the feature statistics of each Transformer block via AdaLN:
\begin{equation}
    \text{AdaLN}(h, c_{\text{vec}}) = \gamma(c_{\text{vec}}) \odot \text{LayerNorm}(h) + \beta(c_{\text{vec}})
\end{equation}
where $h$ is the hidden state, and $\gamma, \beta$ are scale and shift parameters regressed from $c_{\text{vec}}$. This ensures haptic feedback provides a global constraint on the generation process, preventing the ``modality ignoring'' issue common in naive fusion.

\paragraph{Spatio-Temporal Grounding via Cross-Attention.} Visual sequences $c_{\text{seq}}$ are integrated via Multi-Head Cross-Attention (MHCA) to provide spatial grounding. Let $h$ be the intermediate action tokens acting as queries; the keys $K$ and values $V$ are projected from $c_{\text{seq}}$:
\begin{equation}
    Q = h W_Q, \quad K = c_{\text{seq}} W_K, \quad V = c_{\text{seq}} W_V
\end{equation}
\begin{equation}
    \text{MHCA}(h, c_{\text{seq}}) = \text{Softmax}\left(\frac{QK^\top}{\sqrt{d_k}}\right)V
\end{equation}
where $d_k$ is the dimension per attention head and $W_Q, W_K, W_V$ are learnable projection matrices.

\subsection{Training and Inference}
The neural velocity field $v_\theta$ is trained to jointly predict the motion and force components while optimizing the flow-matching objective.

\paragraph{Training Objective with Active Compliance.} The DiT decoding head outputs a hybrid velocity prediction $v_\theta(\mathbf{a}_t^k, k, c_{\text{vec}}, c_{\text{seq}})$. The training objective minimizes the mean squared error against the target velocity $\mathbf{u}_t^k$:
\begin{equation}
    \mathcal{L}_{\text{FM}}(\theta) = \mathbb{E}_{k \sim [0,1], \mathbf{a}_t^0, \mathbf{a}_t^1} \left[ \| v_\theta(\mathbf{a}_t^k, k, c_{\text{vec}}, c_{\text{seq}}) - (\mathbf{a}_t^1 - \mathbf{a}_t^0) \|^2 \right]
\end{equation}
This joint prediction compels the model to internalize the causal relationship between movement and physical feedback, enabling proactive compliance during contact.

\paragraph{Deterministic Inference via ODE Solving.} During inference, actions are generated by solving the Ordinary Differential Equation (ODE) from noise ($k=1$) to data ($k=0$):
\begin{equation}
    d\mathbf{a}_t^k = v_\theta(\mathbf{a}_t^k, k, \mathcal{O}_t) dk
\end{equation}
where $\mathcal{O}_t = \{c_{\text{vec}}, c_{\text{seq}}\}$. By integrating the predicted velocity field using an Euler solver, ForceFlow produces smooth, deterministic trajectories that eliminate the high-frequency jitter common in stochastic diffusion models.

\section{Real-world Experiments Visualization}
\label{app:Visualization}

\begin{figure}[H]
  \centering
  \includegraphics[width=0.9\textwidth]{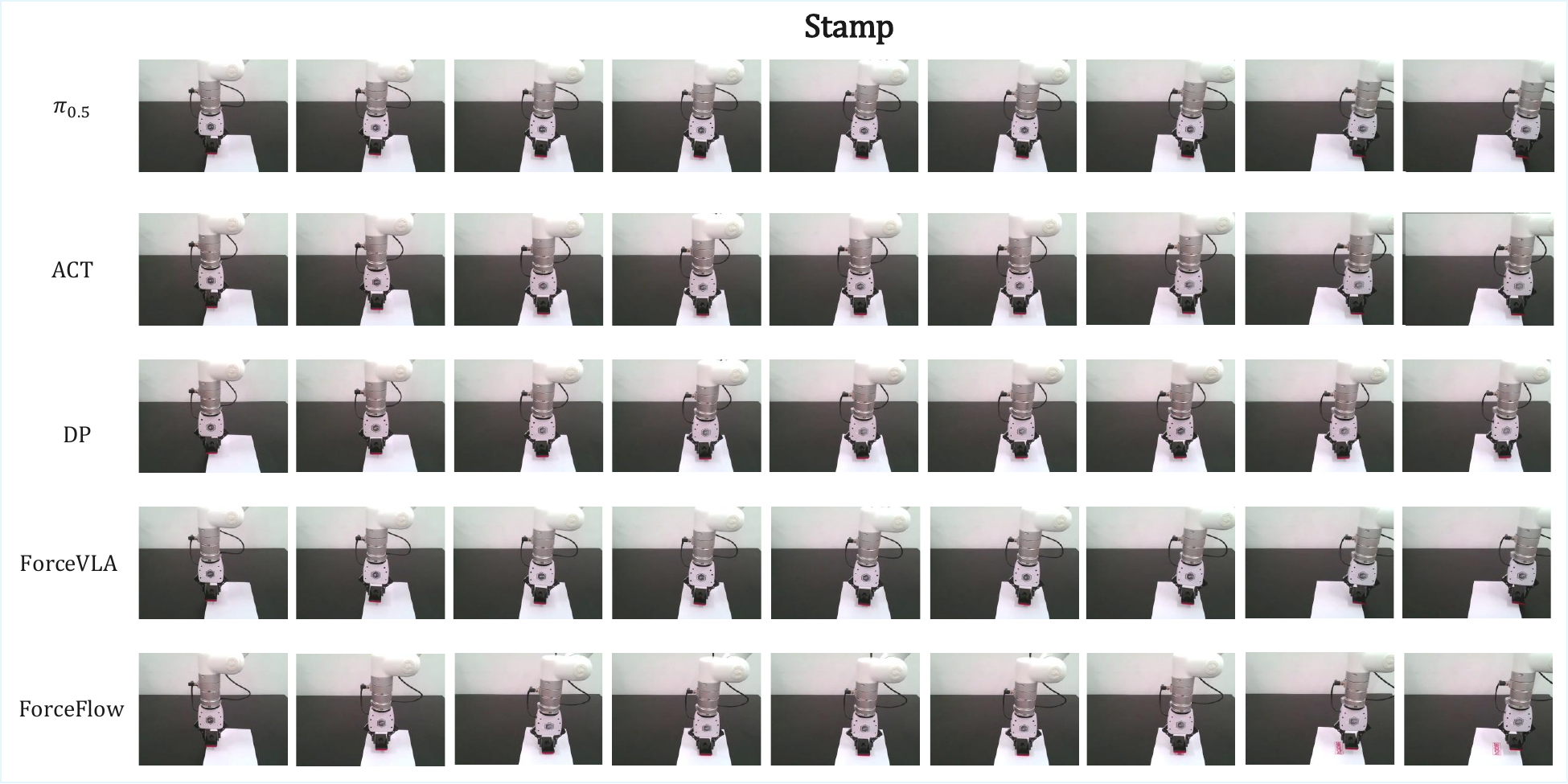}
  \caption{\textbf{Stamp}}
  \label{fig:Stamp}
\end{figure}

\begin{figure}[H]
  \centering
  \includegraphics[width=0.9\textwidth]{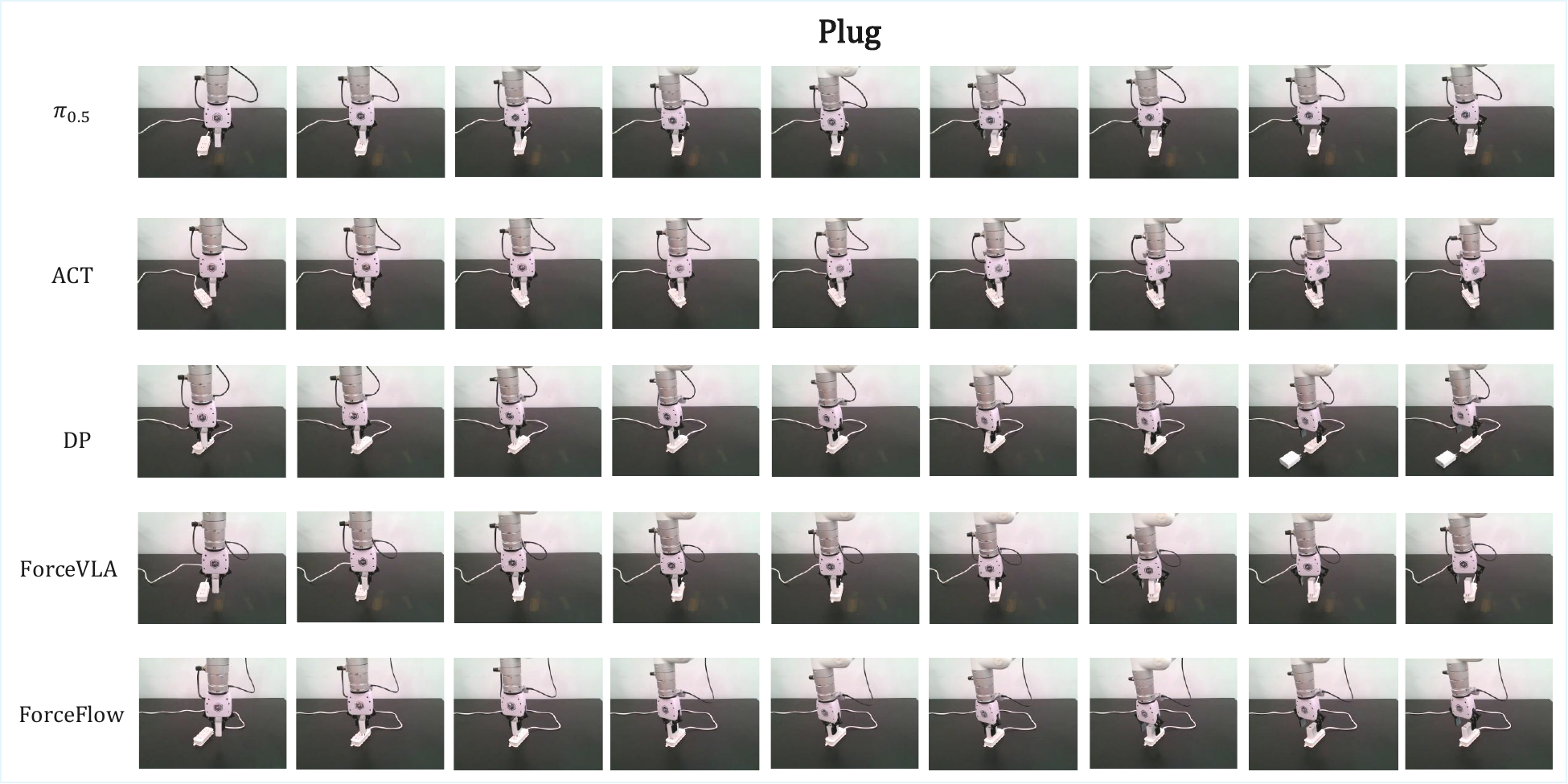}
  \caption{\textbf{Plug}}
  \label{fig:Plug}
\end{figure}

\begin{figure}[H]
  \centering
  \includegraphics[width=0.9\textwidth]{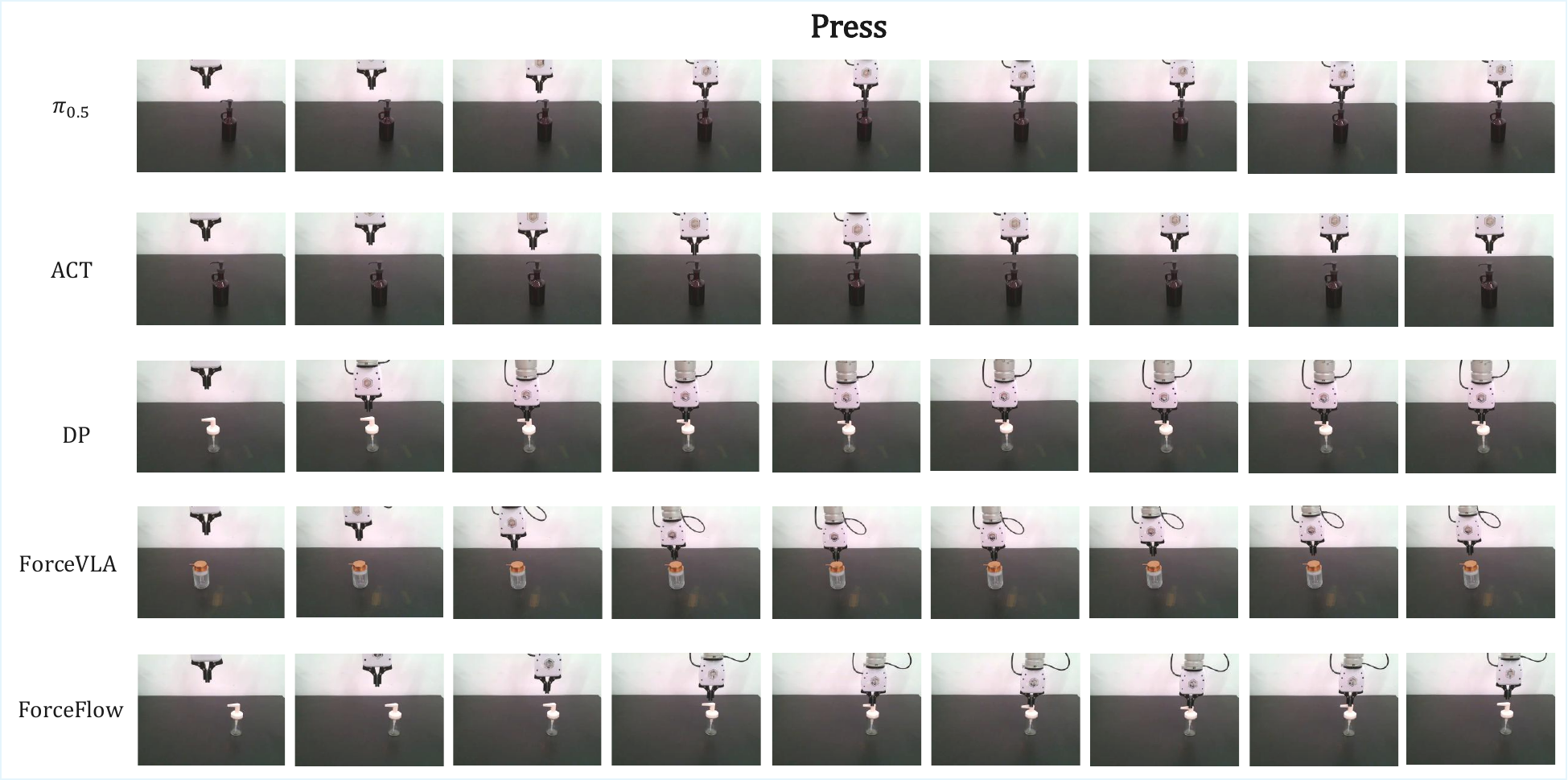}
  \caption{\textbf{Press}}
  \label{fig:Press}
\end{figure}

\begin{figure}[H]
  \centering
  \includegraphics[width=0.9\textwidth]{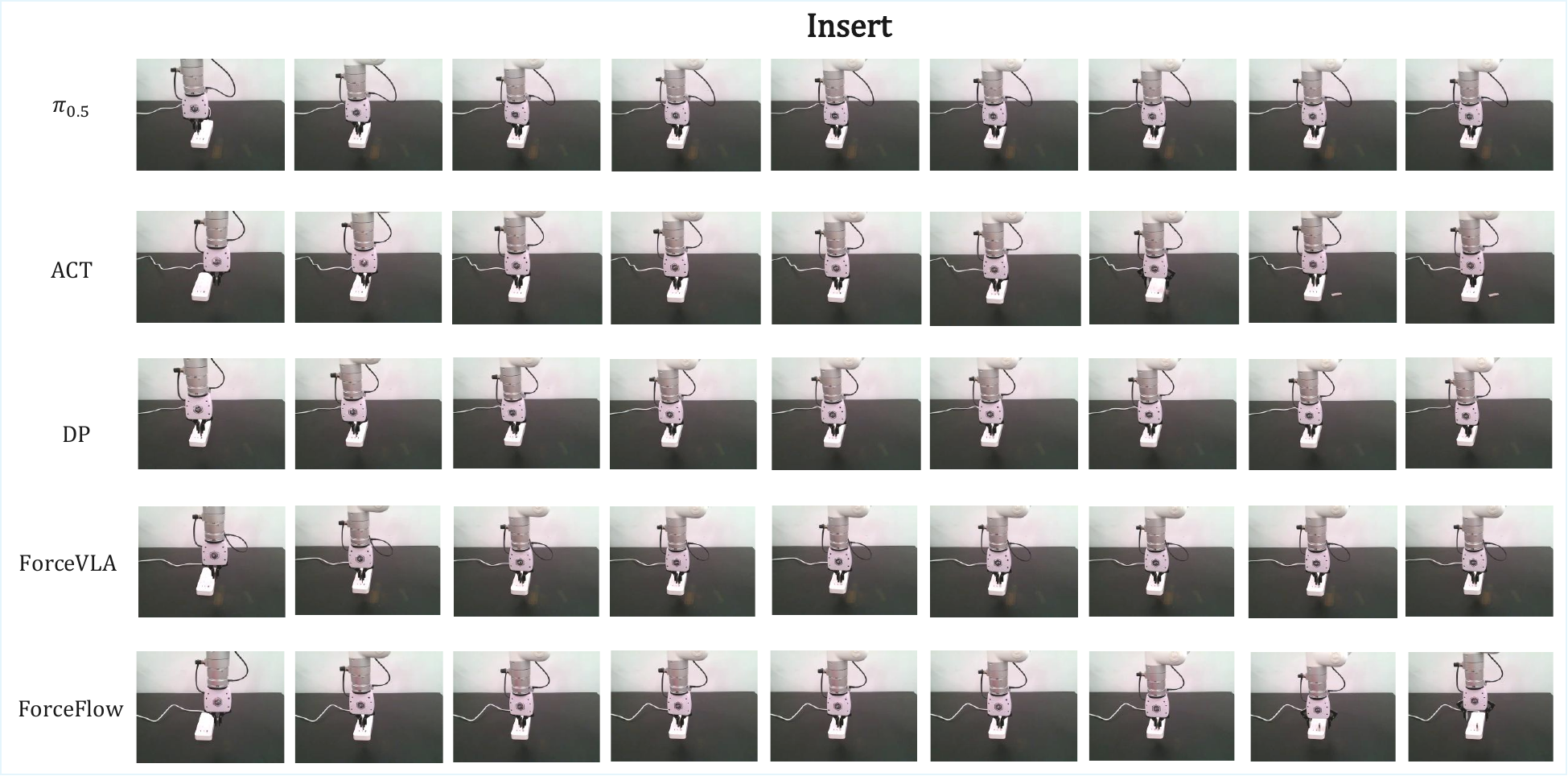}
  \caption{\textbf{Insert}}
  \label{fig:Insert}
\end{figure}

\begin{figure}[H]
  \centering
  \includegraphics[width=0.9\textwidth]{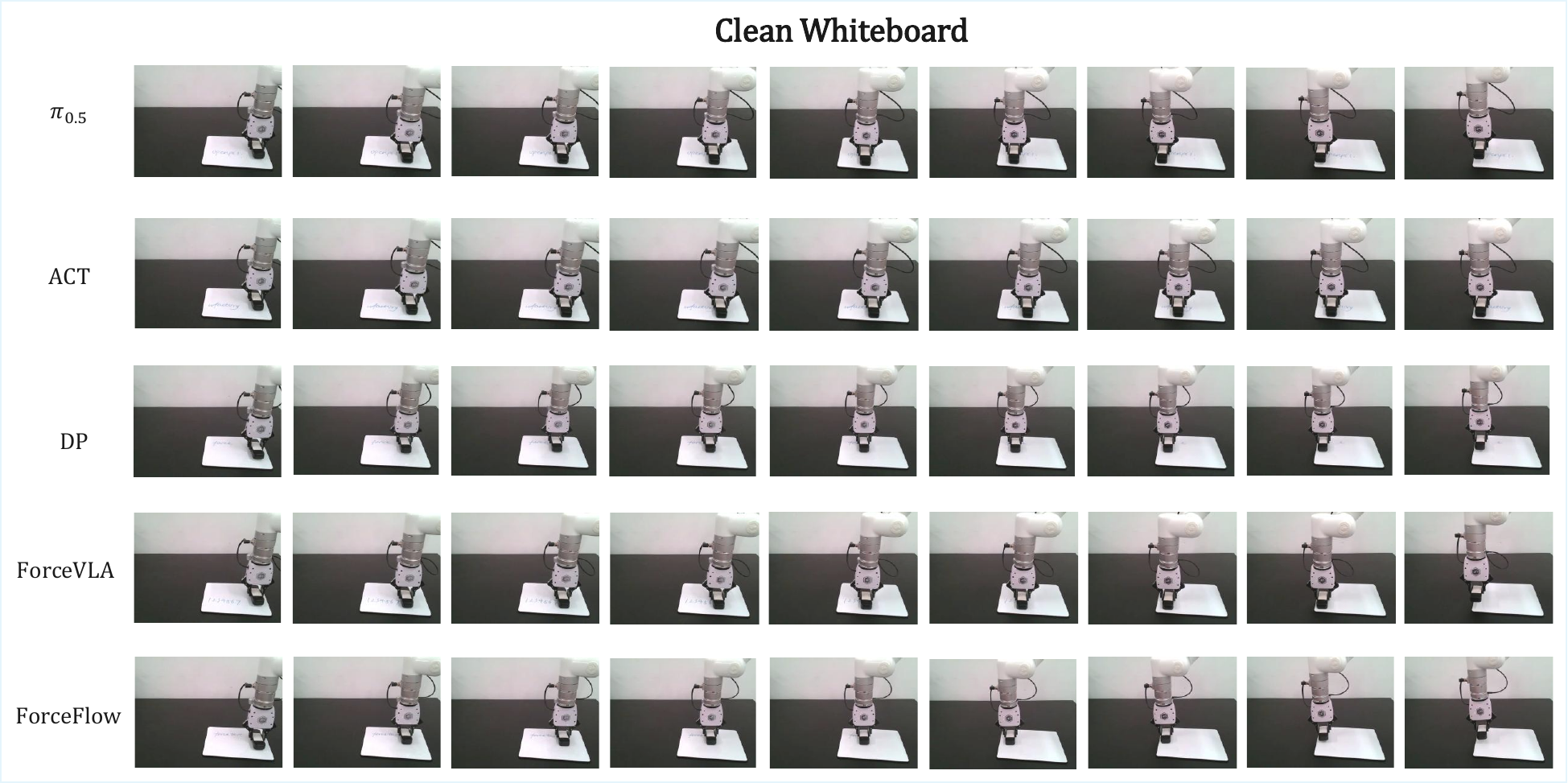}
  \caption{\textbf{Clean Whiteboard}}
  \label{fig:Clean Whiteboard}
\end{figure}

\begin{figure}[H]
  \centering
  \includegraphics[width=0.9\textwidth]{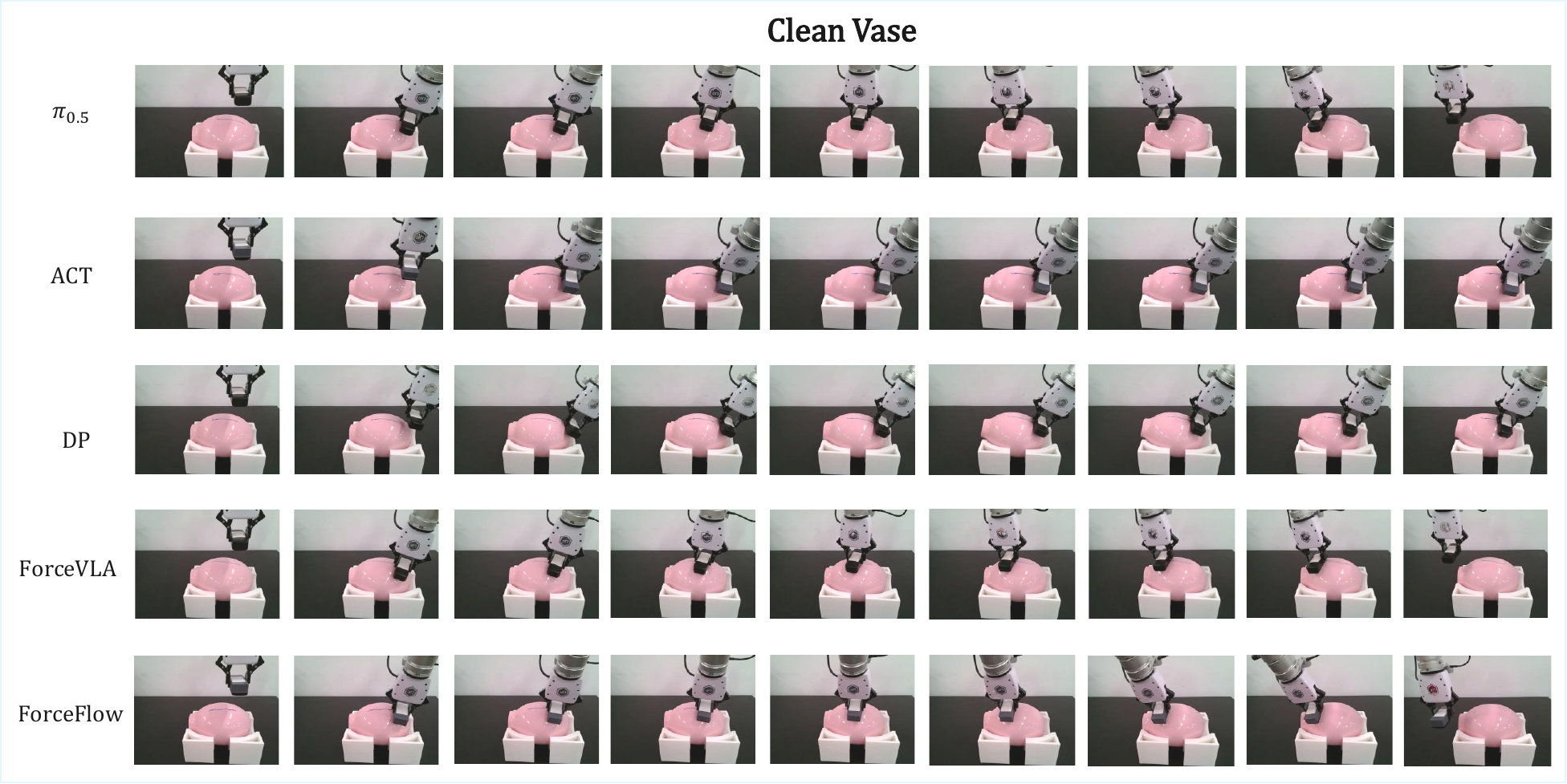}
  \caption{\textbf{Clean vase}}
  \label{fig:Clean vase}
\end{figure}

\section{In-depth Analysis of Force Modalities in Contact-Rich Tasks}
\label{app:force_role}

To further elucidate why \textbf{ForceFlow} significantly outperforms vision-centric baselines, we categorize the role of force perception in our task suite into three distinct functional paradigms based on the interaction physics.

\subsection{Force as the Primary Modality: Resolving Visual Ambiguity}
In scenarios where visual cues are insufficient to determine the state of completion or environmental properties, force perception acts as the primary modality for action grounding.

In \textit{Stamping Task}, the thickness of a paper stack (ranging from 1 to 50 sheets) is visually negligible from a top-down or wrist camera. A vision-only policy tends to converge to a mean terminal height, leading to insufficient pressure on thin stacks or excessive collision on thick ones. ForceFlow utilizes the 10-step force history to detect the exact moment of contact, using the force-torque as stopping signal to trigger the stamp action regardless of the absolute height. In \textit{Press Button Task}, different sanitizer bottles possess varying spring constants and trigger depths. Because the model must reach a specific internal pressure to complete the press, ForceFlow treats the force-torque (F/T) profile as the ground truth for state transition, allowing it to generalize to unseen bottles where pixel-level movement does not directly correlate with successful activation.

\subsection{Force as an Auxiliary Modality: Navigating Geometric Constraints}
In tasks requiring sub-millimeter precision, such as \textit{Plug} and \textit{USB Insertion}, visual alignment provides a coarse initialization, but the successful completion relies on force-guided searching to overcome geometric jamming.

When a USB-A male connector hits the edge of a port, visual feedback often shows the target is reached, but the state is actually a high-friction stall. ForceFlow perceives the reactive torque $\tau$ generated by misalignment. By predicting the target force $\mathbf{F}_{t+1}^{\text{pred}}$, the model generates small-scale lateral sliding and wiggling motions. This force-aware behavior allows the robot to feel the opening and align the components dynamically, a capability entirely absent in vision-only models.

\subsection{Force as a Regulatory Modality: Stable Continuous Interaction}
For tasks involving continuous surface contact, force perception is critical for maintaining a stable normal force ($F_n$), ensuring the end-effector neither loses contact nor damages the substrate.

In the \textit{Clean Whiteboard} task, the normal vector is constant, yet ForceFlow maintains a consistent $5N$ pressure to ensure cleaning efficacy despite potential arm vibration. The \textit{Clean Vase} task introduces non-planar geometry where the normal vector changes continuously. As the arm moves along the curve, ForceFlow adaptively adjusts its joint torques by sensing the change in the F/T vector. The active force prediction mechanism allows the policy to proactively compliance-match the surface curvature, maintaining a stable interaction envelope that vision-centric models fail to track.

\section{Ablation on Visual Modality in Contact Stage}
\label{sec:visual_ablation}

While our primary contribution focuses on alleviating the modality masking issue and proving the necessity of force feedback in contact-rich tasks (as demonstrated by the \textit{w/o Force} baseline in Section 4.2), it is equally important to understand the role of visual input during the interaction stage. The current robotic imitation learning paradigm is inherently vision-centric; the visual modality serves as the absolute foundation for macro-level localization and motion planning. 

To investigate the coupling between vision and force during fine-grained manipulation, we conduct an additional ablation study targeting the visual modality. We design a \textit{w/o vis} variant: the policy is allowed to use visual observations during the VLM-guided \textit{Approach Stage} for spatial localization. However, upon triggering the V2F handover and entering the \textit{Interaction Stage}, we completely mask the visual input, forcing the ForceFlow policy to rely solely on closed-loop force perception and proprioception. We evaluate this variant across our task suite with 10 trials per task.

As shown in Table \ref{tab:vis_ablation}, completely ablating visual input during contact leads to distinct behavioral outcomes depending on the task geometry. For tasks requiring primarily single-axis pressure regulation (e.g., \textit{Stamp}, \textit{Clean Whiteboard}), high-frequency force feedback alone is sufficient to maintain a high success rate (80\% and 90\%, respectively). However, for tasks involving complex spatial trajectories, narrow tolerances, or non-planar geometries (\textit{Plug}, \textit{Insert}, \textit{Clean Vase}), removing visual input leads to complete failure (0\% success rate). This validates our Asymmetric Fusion architecture: both modalities are indispensable. Vision provides continuous spatial grounding to prevent the end-effector from drifting, while force feedback regulates the physical compliance.

\begin{table}[htbp]
    \centering
    \caption{Ablation on Visual Modality in Contact Stage. We report the success rate (\%) over 10 trials for each task. The \textit{w/o vis} variant masks visual input entirely during the interaction stage.}
    \label{tab:vis_ablation}
    \resizebox{0.93\linewidth}{!}{
    \begin{tabular}{lccccccc}
        \toprule
        \textbf{Method} & \textbf{Stamp} & \textbf{Plug} & \textbf{Press} & \textbf{Insert} & \textbf{Clean WB} & \textbf{Clean Vase} & \textbf{Avg.} \\
        \midrule
        ForceFlow (Full) & 85\% & 90\% & 90\% & 60\% & 100\% & 65\% & \textbf{81.6\%} \\
        w/o vis & 80\% & 0\% & 15\% & 0\% & 90\% & 0\% & 30.8\% \\
        \bottomrule
    \end{tabular}
    }
\end{table}

\section{Discussion on Baseline Selection and Observation Modalities}
\label{sec:baseline_discussion}

In Section 4, we primarily compared ForceFlow against the state-of-the-art force-aware model, ForceVLA \citep{forcevla}. This selection was deliberate to ensure a rigorous, controlled evaluation. The current landscape of force-aware imitation learning lacks a unified standard for modality representations, with different frameworks employing vastly different sensor hardware and data structures.

Table \ref{tab:modality_comparison} summarizes the observation modalities utilized by recent related works. Models such as OmniVTLA \citep{omnivtla}, RDP \citep{rdp}, and ViTaL \citep{vital} rely on high-dimensional tactile arrays (e.g., GelSight sensors), which capture fine-grained local geometry but differ significantly in physical properties and data dimensionality from global force-torque measurements. Conversely, TA-VLA \citep{tavla} utilizes internal joint angles and joint torques, while methods like FeelTheForce \citep{feeltheforce} and FoAR \citep{foar} process 3D point clouds rather than 2D RGB images.

ForceVLA is the only state-of-the-art model that shares our exact modality alignment (Multi-view RGB + EEF Pose + 6D EEF Wrench). Comparing models with mismatched observation spaces or underlying hardware platforms confounds the evaluation, making it impossible to determine whether performance gaps stem from the architectural design of the multimodal fusion module or simply from differences in feature extraction and sensor density. By benchmarking against ForceVLA, we strictly control for environmental variables, demonstrating that the significant 37\% performance improvement achieved by ForceFlow is entirely attributable to our proposed flow-matching formulation and the Asymmetric Fusion architecture, which effectively prevents the modality masking issue prevalent in standard MoE or concatenation-based fusion strategies.

\begin{table}[htbp]
    \centering
    \caption{Comparison of Observation Modalities in Force-Aware Imitation Learning Methods. ForceFlow strictly aligns with ForceVLA's input space to ensure a fair architectural comparison.}
    \label{tab:modality_comparison}
    \resizebox{\linewidth}{!}{
    \begin{tabular}{lccc}
        \toprule
        \textbf{Method} & \textbf{Visual Modality} & \textbf{Proprioception} & \textbf{Force/Tactile Modality} \\
        \midrule
        ForceMimic \citep{forcemimic} & 3D Point Clouds & EEF Pose & - \\
        OmniVTLA \citep{omnivtla} & Multi-view RGB & EEF Pose & High-dim Tactile Arrays \\
        RDP \citep{rdp} & RGB & EEF Pose & 3D Tactile Deformation Field \\
        ViTaL \citep{vital} & RGB & EEF Pose & High-dim Tactile Arrays \\
        FeelTheForce \citep{feeltheforce} & 3D Key points & 3D Key points & 1D Norm Force\\
        FoAR \citep{foar} & 3D Point Clouds & EEF Pose & 6D EEF Wrench \\
        TA-VLA \citep{tavla} & Multi-view RGB & Joint & Joint Torques \\
        \midrule
        ForceVLA \citep{forcevla} & Multi-view RGB & EEF Pose & 6D EEF Wrench \\
        \textbf{ForceFlow (Ours)} & Multi-view RGB & EEF Pose & 6D EEF Wrench (w/ History) \\
        \bottomrule
    \end{tabular}
    }
\end{table}

\section{Additional Experiments and System Details}
\label{sec:additional_experiments}

In this section, we provide supplementary experiments and implementation details to further evaluate the fairness, efficiency, and hardware reliability of the ForceFlow framework.

\subsection{Fairness of the V2F Mechanism in OOD Scenarios}
To further validate the superiority of our Hierarchically Decoupled architecture (V2F) in handling spatial Out-of-Distribution (OOD) scenarios, we conducted an additional evaluation equipping all baseline methods with the identical V2F upper-level module. 

As shown in Table~\ref{tab:v2f_fairness}, even with the V2F module solving the coarse navigation and spatial alignment problem, the end-to-end baseline policies still fail drastically in contact-rich execution. This demonstrates that once spatial alignment is achieved, the physical interaction and force regulation become the true bottleneck. The significant performance gap confirms that the V2F mechanism does not merely provide an "unfair structural advantage" in navigation, but rather enables the lower-level ForceFlow policy to fundamentally resolve complex contact dynamics.

\begin{table}[h]
\centering
\caption{Spatial OOD Generalization with V2F Equipped Baselines (SR \%). We conducted 10 independent trials per task to ensure a fair comparison.}
\label{tab:v2f_fairness}
\begin{tabular}{lcccc}
\toprule
Method & Press & Plug & Clean WB & Avg. Success \\
\midrule
$\pi_{0.5}$ + V2F & 0\% & 0\% & 0\% & 0\% \\
ACT + V2F & 0\% & 0\% & 0\% & 0\% \\
Diffusion Policy + V2F & 0\% & 0\% & 0\% & 0\% \\
ForceVLA + V2F & 20\% & 0\% & 20\% & 13.33\% \\
\midrule
\textbf{ForceFlow + V2F (Ours)} & \textbf{40\%} & \textbf{10\%} & \textbf{50\%} & \textbf{33.33\%} \\
\bottomrule
\end{tabular}
\end{table}

\subsection{System Latency and Inference Frequency}
To ensure a rigorous comparison of system efficiency, we evaluated the inference latency and execution details across all methods. The underlying robot control frequency is uniformly fixed at 30 Hz across all baselines to maintain physical execution consistency. 

For the proposed framework, the VLM in the V2F module runs only once to provide a coarse target. ForceFlow then takes over for real-time continuous control. At each timestep, the policy predicts a 64-step action chunk. By executing only the first 32 steps ($\sim 1$s) and immediately updating all sensory inputs for the next inference, the system forms a continuous closed loop without excessive delay. Detailed latency metrics are provided in Table~\ref{tab:latency}.

\begin{table}[h]
\centering
\caption{Comparison of System Latency and Execution Details.}
\label{tab:latency}
\resizebox{\columnwidth}{!}{%
\begin{tabular}{lcccc}
\toprule
Method & Inference Latency & Action Horizon & Executed Steps & Replanning Cycle \\
\midrule
ACT & $\sim 6.4$ ms & 100 & 100 & $\sim 3.340$ s \\
$\pi_{0.5}$ & $\sim 81.3$ ms & 50 & 25 & $\sim 0.915$ s \\
Diffusion Policy & $\sim 109.9$ ms & 16 & 8 & $\sim 0.377$ s \\
ForceVLA & $\sim 84.1$ ms & 50 & 25 & $\sim 0.917$ s \\
\midrule
V2F (Upper-level) & $\sim 1805$ ms & - & - & Runs only once \\
\textbf{ForceFlow (Lower-level)} & $\sim 83.3$ ms & 64 & 32 & $\sim 1.150$ s \\
\bottomrule
\end{tabular}%
}
\end{table}

\subsection{Data Collection Interface and Hardware Setup}
\begin{figure}[H]
  \centering
  \includegraphics[width=0.9\textwidth]{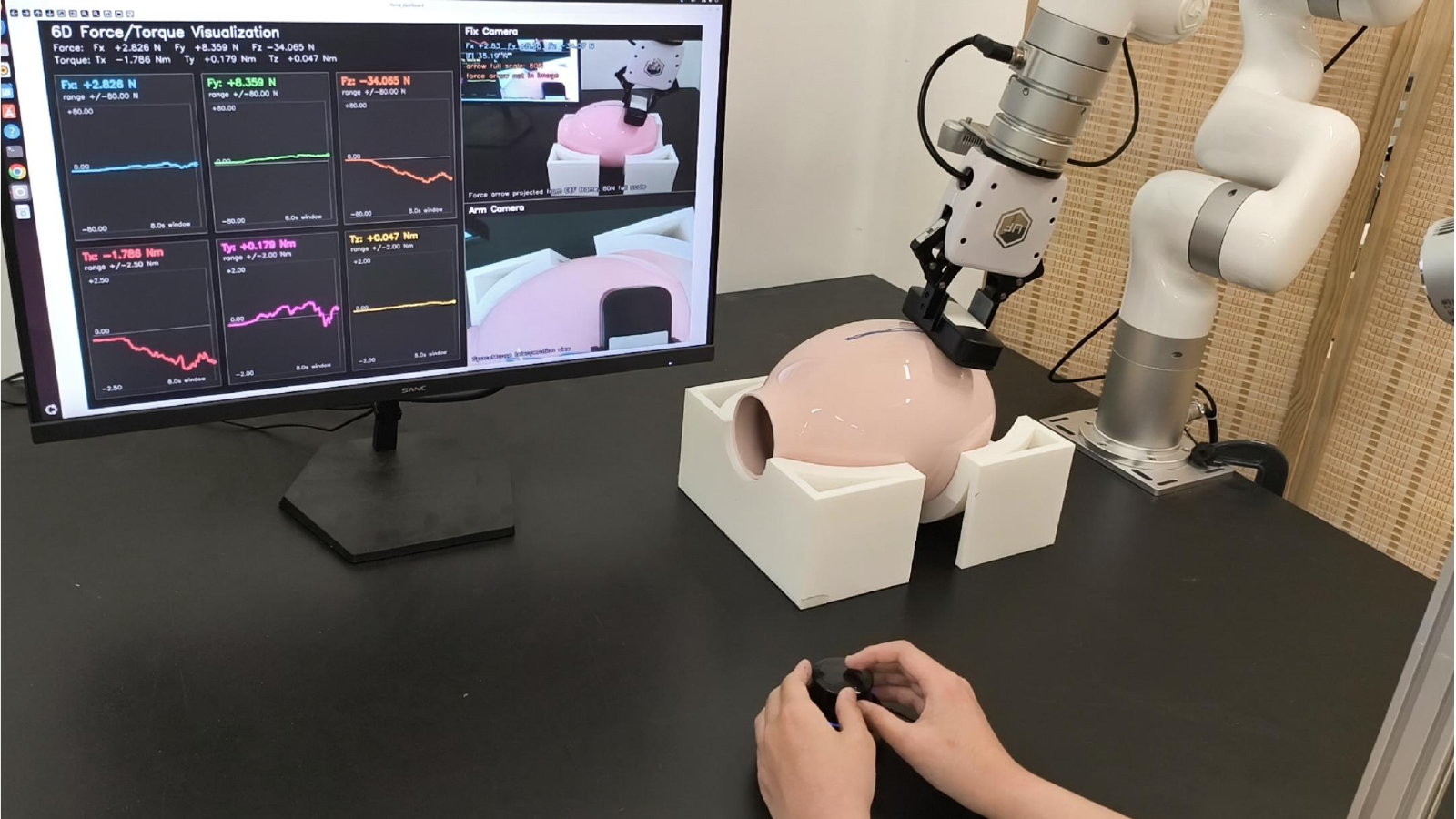}
  \caption{\textbf{Custom Real-time Force Visualization UI.} The interface displays synchronized multi-view visual feeds (Arm Camera and Fix Camera) alongside real-time 6D force ($F_x, F_y, F_z$) and torque ($T_x, T_y, T_z$) dynamic curves, enabling operators to achieve precise visuo-motor compensation during teleoperation.}
  \label{fig:app_ui}
\end{figure}

Our data collection pipeline is designed to ensure high-quality, physically grounded expert demonstrations. To avoid the prohibitive hardware costs associated with active haptic feedback systems, we developed a Custom Real-time Force Visualization UI (see Figure~\ref{fig:app_ui}).

During teleoperation, operators monitor precise force/torque numerical panels and dynamic curves alongside the multi-view visual feeds. This explicit feedback loop allows operators to perform precise visuo-motor compensation during complex contact phases, ensuring the recorded demonstrations are deliberate, safe, and of high quality. Furthermore, to capture reliable and high-fidelity interaction forces, our hardware system utilizes the official UFACTORY xArm 6-axis force torque sensor equipped on the robot's end-effector.

\end{document}